\documentclass{article} % For LaTeX2e
\usepackage{tvmcn_preprint_base,times}

\usepackage{amsmath,amsfonts,bm}

\def\eqref#1{equation~\ref{#1}}
\def\1{\bm{1}}

\DeclareMathAlphabet{\mathsfit}{\encodingdefault}{\sfdefault}{m}{sl}
\SetMathAlphabet{\mathsfit}{bold}{\encodingdefault}{\sfdefault}{bx}{n}

\usepackage{amsmath,amssymb}
\usepackage{booktabs}
\usepackage{multirow}
\usepackage{graphicx}
\usepackage{flafter}
\usepackage{placeins}
\usepackage{hyperref}
\usepackage{url}
\usepackage{array}
\usepackage[normalem]{ulem}
\usepackage{makecell}
\usepackage{tabularx}
\usepackage{caption}
\usepackage{tvmcn_arxiv}

\title{A Time-Variant Spiking Neuron from a Unified State-Dynamics Perspective}

\author{%
\begin{minipage}[t]{\dimexpr\textwidth-2\tabcolsep\relax}
\centering
\normalfont
\textbf{Jing Yang$^{1}$ \quad Qianhui Liu$^{1}$ \quad
Gang Pan$^{2}$ \quad Zhanglu Yan$^{3}$}\\[6pt]
{\small
$^{1}$School of Artificial Intelligence, Shandong University, Jinan, China\\
$^{2}$College of Computer Science and Technology, Zhejiang University, Hangzhou, China\\
$^{3}$Department of Computer Science, National University of Singapore, Singapore\\[4pt]
\texttt{qhliu@sdu.edu.cn}
}
\end{minipage}%
}


\begin{document}

\maketitle

\begin{abstract}
Spiking neural networks (SNNs) have increasingly incorporated rich neuronal dynamics to enhance temporal processing. However, richer dynamics do not necessarily enable neurons to adapt how information is processed over time. To examine this issue systematically, we introduce a unified state-dynamics framework that characterizes diverse spiking neurons through a common formulation and enables direct comparison of their dynamical designs. The framework reveals a common limitation of existing neurons: although they enhance temporal processing through richer internal states and state evolution, the dynamical rules determining how inputs are integrated, how states evolve, and how they are read out generally remain unchanged throughout temporal processing. Such time-invariant rules limit the ability of neurons to adjust information processing as information relevance changes over time. Inspired by time-varying regulation in biological neurons, we propose the Time-Variant Multi-Compartment Neuron (TV-MCN). TV-MCN builds on multiscale compartment dynamics and makes the rules for input integration, state evolution, and state readout time-varying according to the current input. Specifically, an input-dependent internal time regulates the rate of state evolution, while the current input dynamically modulates input integration and state readout. Extensive experiments show that TV-MCN achieves strong overall performance, with particularly large gains on tasks requiring selective retention and retrieval of information, while ablation studies further support the effectiveness of its time-varying mechanisms.
\end{abstract}

\section{Introduction}
Spiking neural networks (SNNs) are biologically inspired models that communicate through discrete spikes between neurons. Within each neuron, internal states integrate incoming signals and evolve over time, determining how past inputs influence future responses. The design of these state dynamics is therefore central to the temporal processing capability of SNNs. Recent studies have extended neuronal dynamics by introducing heterogeneous time constants to capture multiple retention scales \citep{zheng2024temporal,deng2025hetsyn}, interacting compartments for modeling structured internal dynamics \citep{zhang2024tclif,chen2026pmsn}, and incorporating resonant modes with distinct frequency responses  \citep{zhang2025drf}. These developments enrich the temporal dynamics of spiking neurons beyond the single decay mode of the leaky integrate-and-fire (LIF) neuron.

However, richer dynamics do not necessarily enable neurons to adapt how information is processed over time.
In real-world settings, useful information is often interleaved with irrelevant distractors, and information that is important at one moment may become less relevant as new inputs arrive. Neurons therefore are required to determine what information to integrate or discard, what to maintain or forget, and which stored information should be used for the output \citep{chatham2015multiple,unger2016working}. Yet it remains unclear whether existing neuron designs enable such flexible information processing.

To examine this question systematically, we propose a unified state-dynamics framework that describes spiking-neuron computation through three biologically motivated stages: input integration and state evolution, state readout to form the somatic membrane potential, and spike generation followed by post-spike state recovery \citep{hodgkin1952quantitative,magee2000dendritic,bean2007action}. Neuron dynamics developed from different biological motivations and mathematical formulations can be expressed as different parameterizations of this framework, enabling their dynamical structures and design choices to be directly compared and analyzed within a common formulation.

Using this unified framework, we identify a common limitation of existing designs. Although they enhance temporal processing by increasing the dimensionality or diversifying the temporal evolution of internal states, their dynamical rules determining how inputs are integrated into these states, how the states evolve, and how they are read out generally remain unchanged throughout temporal processing. Consequently, such time-invariant rules limit the ability of neurons to adjust how information is integrated, maintained, and used as the relevance of incoming information changes.

Inspired by the time-varying regulation observed in biological neurons \citep{schulz2006cellular}, we propose the Time-Variant Multi-Compartment Neuron (TV-MCN). TV-MCN builds on multiscale compartment dynamics that provide diverse intrinsic temporal scales for representing information. On this basis, it makes the dynamical rules for input integration, state evolution, and state readout time-varying according to the current input. Specifically, an input-dependent internal time regulates the rate of state evolution, while the current input dynamically modulates input integration and state readout. Together, these mechanisms enable TV-MCN to adjust how information is integrated, maintained, and used as information relevance changes over time.

We evaluate TV-MCN across a broad range of temporal classification and synthetic sequence benchmarks. The classification benchmarks include Google Speech Commands (GSC), Spiking Heidelberg Digits (SHD) and Spiking Speech Commands (SSC), and DVS128 Gesture \citep{warden2018speech,cramer2022heidelberg,amir2017lowpower}, while the synthetic tasks include Adding, Selective Copying, 20D STORE--RECALL, MAD ICR and MAD Noisy ICR, and Induction \citep{hochreiter1997long, salaj2021spike, poli2024mechanistic, gu2023mamba}. TV-MCN achieves strong performance across the benchmarks and substantially outperforms existing spiking neuron models on tasks that require selective retention and retrieval of information. For example, on MAD Noisy ICR \citep{poli2024mechanistic}, TV-MCN reaches an accuracy of 99.99\%, compared with 85.43\% for HetSyn \citep{deng2025hetsyn} and 40.15\% for D-RF \citep{zhang2025drf}. Further ablation studies confirm the contribution of its time-varying dynamics. Our contributions are summarized as follows:
\begin{itemize}
    \item We introduce a unified state-dynamics framework that characterizes diverse spiking neurons through a common formulation, revealing that the governing dynamical rules are temporally unchanged in existing designs.
    \item We propose the Time-Variant Multi-Compartment Neuron (TV-MCN), which introduces explicit time variation into input integration, state evolution, and state readout, enabling adaptive information processing as inputs change over time.
    \item Experimental results demonstrate that TV-MCN achieves strong overall performance, with particularly large gains on tasks requiring selective retention and retrieval, validating the effectiveness of its time-varying dynamics.
\end{itemize}

\section{A Unified Framework for Neuronal Dynamics}
In biological neurons, membrane potentials, synaptic conductances, and ion-channel states continuously evolve over time, while synaptic inputs drive these internal states by changing local currents and conductances \citep{hodgkin1952quantitative}. The electrical signals generated by these states propagate and integrate within dendrites and jointly influence the somatic membrane potential \citep{magee2000dendritic}. Once the firing condition is reached, the neuron generates a spike, followed by post-spike recovery that updates its internal states \citep{bean2007action}. The biological facts above can be summarized as follows: neuronal internal states are driven by current inputs and evolve over time; these states jointly influence the membrane potential, which in turn determines spike generation. 

\begin{figure}[!t]
\centering
\includegraphics[width=0.85\linewidth]{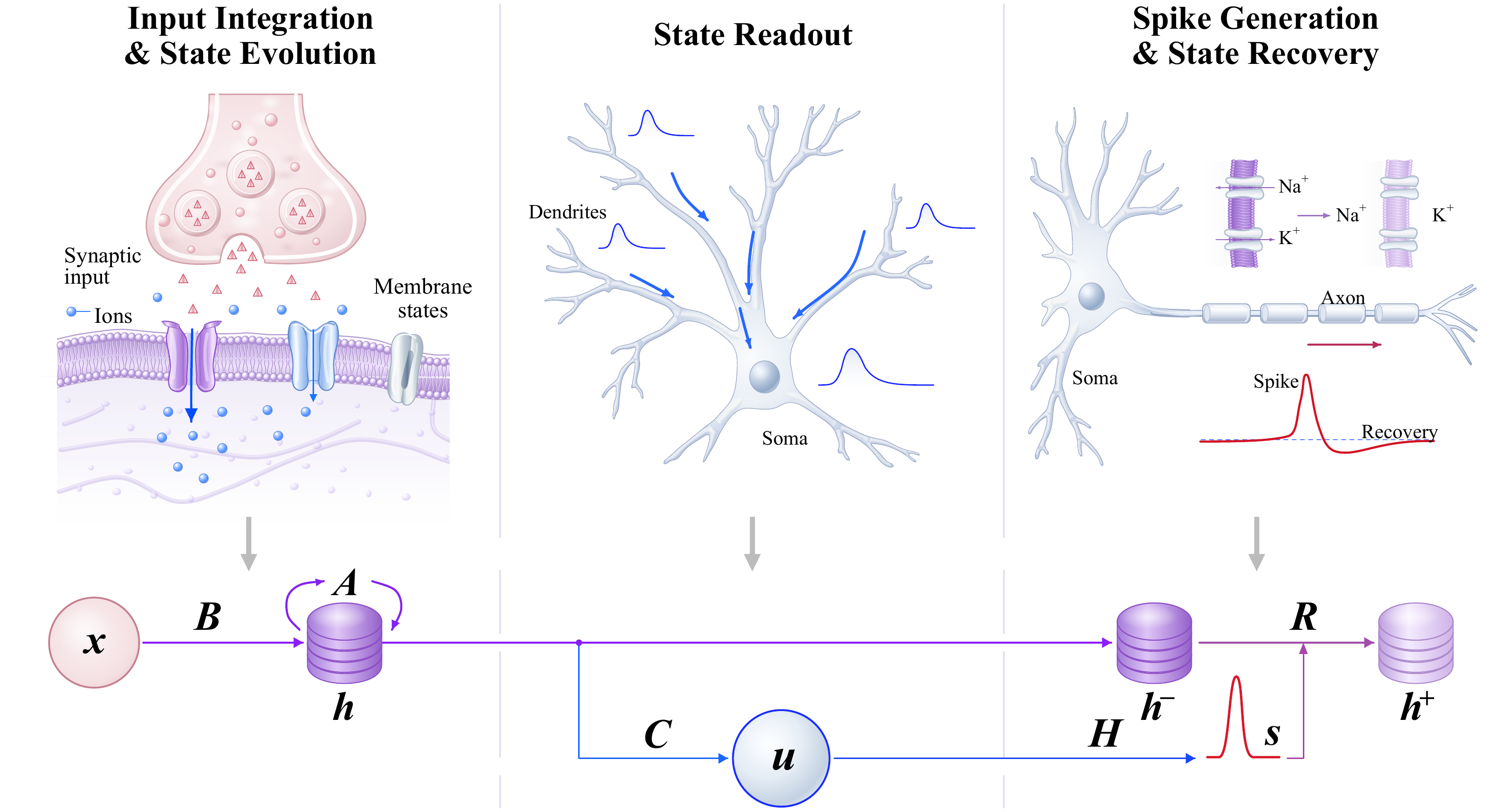}
\caption{From biological neuronal computation to the unified state-dynamics framework. Left: Synaptic inputs drive internal states, corresponding to input integration through $B$ and state evolution through $A$. Middle: Dendritic states are read out at the soma to form the membrane potential. Right: Spike generation is followed by ionic currents that drive post-spike state recovery.}
\label{fig:unified_framework}
\end{figure}

Based on this biological process, we abstract the spiking neuron as a state-dynamics model in which the internal states carry information and evolve under incoming inputs. As illustrated in Figure~\ref{fig:unified_framework}, neuronal computation is decomposed into three stages: (1) input integration and state evolution, (2) state readout, and (3) spike generation and state recovery. The corresponding dynamics can be written as:
\begin{equation}
\begin{aligned}
\dot h(t) &= A h(t)+B x(t),\\
u(t) &= C h(t),\\
s(t) &= H\!\left(u(t)-\vartheta(t)\right),\qquad h(t^{+}) = R\!\left(h(t^{-}),s(t)\right).
\end{aligned}
\end{equation}
where $h(t)$ denotes the internal states, $x(t)$ the input, and $u(t)$ the membrane potential. In the first stage, $A$ governs how these states evolve, including decay and oscillation, while $B$ determines how the current input is integrated into states. In the second stage, $C$ specifies how the internal states are read out to form membrane potential. In the final stage, $H$ denotes spike generation and $R$ describes the post-spike state recovery. $h(t^-)$ and $h(t^+)$ denote the states immediately before and after the recovery, respectively. 

\begin{table}[!b]
\centering
\caption{Parameterizations of spiking neurons under the unified state-dynamics framework.}
\label{tab:neuron_parameterization}
\footnotesize
\setlength{\tabcolsep}{2.0pt}
\renewcommand{\arraystretch}{1.45}
\scalebox{0.9}{
\begin{tabular*}{\linewidth}{@{\extracolsep{\fill}}lllll@{}}
\toprule
Neuron & $h$ &  $A$ &  $B$ &  $C$ \\
\midrule
LIF \citep{burkitt2006review}
& $u$
& $-\tau_m^{-1}$
& $W\tau_m^{-1}$
& $1$ \\

ad-LIF \citep{baronig2025advancing}
& $[u,w]^{\mathsf T}$
& $\left[\begin{smallmatrix}-\tau_u^{-1}&-\tau_u^{-1}\\a\tau_w^{-1}&-\tau_w^{-1}\end{smallmatrix}\right]$
& $\operatorname{diag}(\tau_u^{-1},b\tau_w^{-1})$
& $e_1^{\mathsf T}$ \\

TC-LIF \citep{zhang2024tclif}
& $[U^D,U^S]^{\mathsf T}$
& $\left[\begin{smallmatrix}1&\beta_1\\\beta_2&1+\beta_1\beta_2\end{smallmatrix}\right]$
& $\left[\begin{smallmatrix}1&-\gamma\\\beta_2&-(\beta_2\gamma+V_{\rm th})\end{smallmatrix}\right]$
& $e_2^{\mathsf T}$ \\

DGN \citep{bai2025gating}
& $[\mathbf D^{\mathsf T},V]^{\mathsf T}$
& $\left[\begin{smallmatrix}-\tau_s^{-1}I_N&0\\\mathbf w^{\mathsf T}&-(g_l+\mathbf c^{\mathsf T}\mathbf D)\end{smallmatrix}\right]$
& $\left[\begin{smallmatrix}\tau_s^{-1}I_N\\0\end{smallmatrix}\right]$
& $e_{N+1}^{\mathsf T}$ \\

DH-LIF \citep{zheng2024temporal}
& $[\mathbf i^{\mathsf T},u]^{\mathsf T}$
& $\left[\begin{smallmatrix}\Lambda_\alpha&0\\(1-\beta)\boldsymbol\alpha^{\mathsf T}&\beta\end{smallmatrix}\right]$
& $\left[\begin{smallmatrix}I_D-\Lambda_\alpha\\(1-\beta)(\mathbf1-\boldsymbol\alpha)^{\mathsf T}\end{smallmatrix}\right]$
& $e_{D+1}^{\mathsf T}$ \\

HetSyn \citep{deng2025hetsyn}
& $[\mathbf I^{\mathsf T},J]^{\mathsf T}$
& $-\operatorname{diag}(\tau_{s,1}^{-1},\ldots,\tau_{s,M}^{-1},\tau_J^{-1})$
& $\operatorname{diag}(w_1,\ldots,w_M,\vartheta)$
& $[\mathbf1_M^{\mathsf T},-1]$ \\

D-RF \citep{zhang2025drf}
& $[z_i]_{i=1}^{n}$
& $\operatorname{diag}(-\tau_i^{-1}+j\omega_i)$
& $\boldsymbol\gamma$
& $\mathbf c^{\mathsf T}$ \\

PMSN \citep{chen2026pmsn}
& $[v_i]_{i=1}^{n}$
& $-\operatorname{diag}(\tau_i^{-1})+A_\beta$
& $\boldsymbol\gamma$
& $e_n^{\mathsf T}$ \\
\bottomrule
\end{tabular*}}

\vspace{2pt}
\parbox{\linewidth}{\footnotesize 
Here, \(\tau\) denotes time constants, \(\omega\) oscillation frequencies, and \(A_\beta\) inter-compartment coupling. \(W\), \(\boldsymbol{\gamma}\), and related coefficients specify input integration; \(e_k^{\mathsf T}\) selects the \(k\)-th state, and \(\mathbf c^{\mathsf T}\) defines a weighted state readout. Other symbols denote model-specific internal states and parameters as defined in the cited works.}
\end{table}

% Under this framework, spiking neurons with different biological motivations and mathematical forms can be viewed as different parameterizations of the internal state $h$, state evolution $A$, input integration $B$, and state readout $C$, as summarized in Table~\ref{tab:neuron_parameterization}. This unified formulation allows their dynamical structures and design choices to be directly compared and analyzed. It reveals that additional compartments and synaptic-current states both expand the dimensionality of $h$, while heterogeneous time constants and state coupling shape the decay and oscillatory modes represented by $A$. Differences in input integration and state readout are reflected in how \(B\) distributes inputs across these states and how \(C\) reads out a single state or a weighted combination of states. Further analyses of representative neuron models under this framework are provided in the Appendix.

Under this framework, spiking neurons with different biological motivations and mathematical forms can be viewed as different parameterizations of the internal state $h$, state evolution $A$, input integration $B$, and state readout $C$, as summarized in Table~\ref{tab:neuron_parameterization}. This unified formulation allows their dynamical structures and design choices to be directly compared and analyzed. It reveals that additional compartments \citep{zhang2024tclif,chen2026pmsn} and synaptic-current states \citep{deng2025hetsyn} expand the dimensionality of $h$, while heterogeneous time constants \citep{zheng2024temporal,deng2025hetsyn}, state coupling \citep{zhang2024tclif,chen2026pmsn}, and resonant dynamics \citep{zhang2025drf} enrich the state evolution expressed by the structure of $A$. Differences in input integration and state readout are reflected in how \(B\) distributes inputs across these states and how \(C\) reads out either a single state or a weighted combination of states \citep{chen2026pmsn,zhang2025drf}. Further analyses of representative neuron models under this framework are provided in Appendix~\ref{app:framework_formulations}.

The framework further highlights a key distinction between biological neurons and existing spiking neuron designs. Biological neuronal dynamics are continuously regulated during information processing, whereas existing designs generally rely on fixed dynamical rules. In biological neurons, these dynamical processes are regulated through multiple mechanisms. Membrane conductance changes affect the effective membrane time constant and state-evolution rate \citep{cavallari2014comparison}, NMDA receptors modulate input integration \citep{li2021nmda}, and dendritic or somatic inhibition regulates state contributions to the somatic response \citep{jadi2012location}. By contrast, most of the representative spiking neurons above use coefficients \(A\), \(B\), and \(C\) that remain unchanged, leaving input integration, state evolution, and state readout without time-variant regulation.

\begingroup
\setlength{\abovedisplayskip}{4pt plus 2pt minus 2pt}
\setlength{\belowdisplayskip}{4pt plus 2pt minus 2pt}
\setlength{\abovedisplayshortskip}{2pt plus 2pt minus 2pt}
\setlength{\belowdisplayshortskip}{2pt plus 2pt minus 2pt}

\section{Time-Variant Multi-Compartment Neuron}
Motivated by the dynamic regulation observed in biological neurons, we redesign neuronal dynamics within the unified framework and propose the Time-Variant Multi-Compartment Neuron (TV-MCN). TV-MCN builds on multiple compartments with distinct decay rates and oscillation frequencies. On this basis, we introduce time-variant regulation by making the dynamical rules depend on the current input. Specifically, internal time regulates the rate of state evolution, while $B_t$ and $C_t$ regulate input integration and state readout, respectively.

\begin{figure}[!htbp]
\centering
\includegraphics[width=0.85\linewidth]{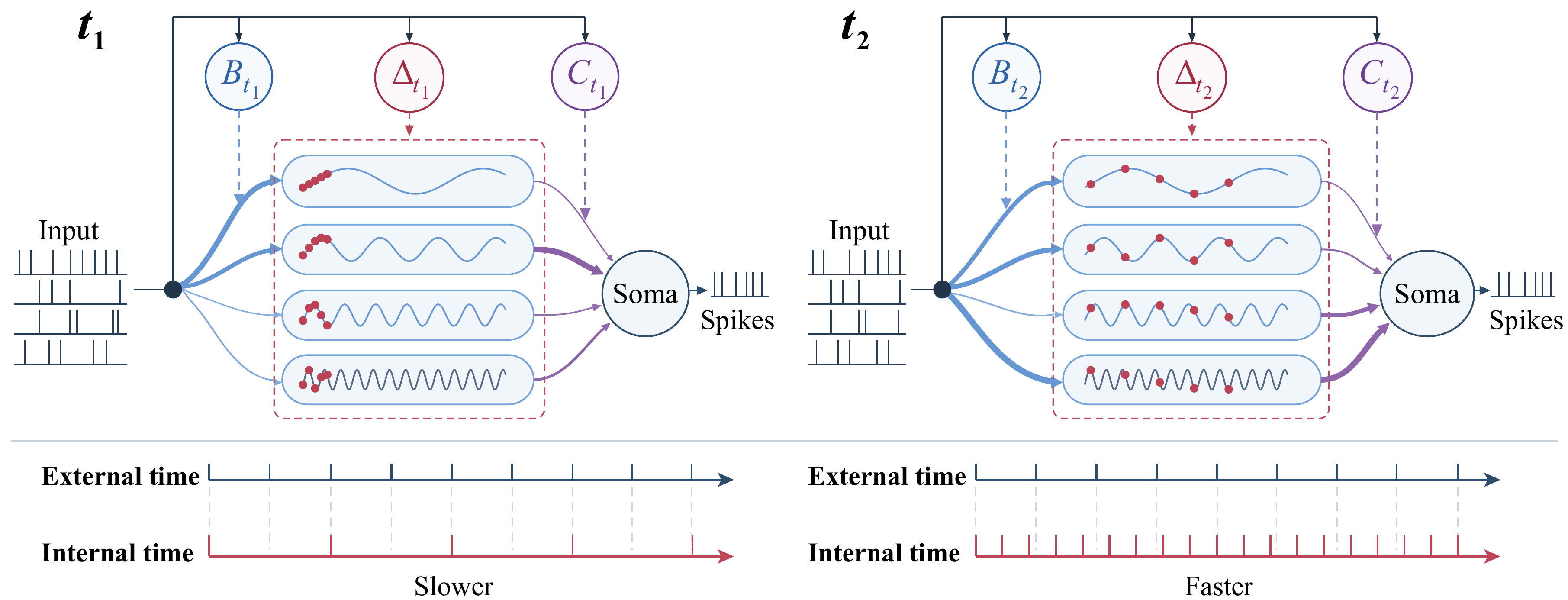}
\caption{Input-dependent dynamics of TV-MCN. At $t_1$ (left), the smaller $\Delta_{t_1}$ slows internal time (bottom axis) and state evolution (red dots on blue curves at equal external-time intervals). At $t_2$ (right), the larger $\Delta_{t_2}$ accelerates both. Different inputs at these moments generate different $B_t$ and $C_t$; blue and purple arrow widths indicate input-integration and state-readout strengths, respectively.}
\label{fig:tvmcn_overview}
\end{figure}

In the conventional formulation, internal states evolve directly with external time $t$:
\begin{equation}
\frac{\mathrm{d} h}{\mathrm{d} t}=A h+B x .
\end{equation}
Here, neuronal dynamics evolve at a fixed rate as external time advances. However, different inputs may require different rates of state evolution. Irrelevant information can be attenuated more rapidly by accelerating state evolution, whereas useful information can be preserved for longer by slowing the rate down.
To enable such adaptive temporal progression, we introduce an internal time $\tau$ and formulate the state dynamics with respect to it:
\begin{equation}
\frac{\mathrm{d} h}{\mathrm{d} \tau}=A h+B_t x .
\end{equation}
We then introduce a progression coefficient $\Delta_t$ that determines how fast internal time advances relative to external time:
\begin{equation}
\frac{\mathrm{d} \tau}{\mathrm{d} t}=\Delta_t, \quad \Delta_t>0.
\end{equation}
Accordingly, over an external-time interval $\mathrm{d}t$, internal time advances by
\begin{equation}
\mathrm{d}\tau=\Delta_t\,\mathrm{d}t.
\end{equation}
A small $\Delta_t$ slows the progression of internal time and thus the evolution of internal states, allowing information to be preserved for longer. In contrast, a large $\Delta_t$ accelerates state evolution, allowing stored information to change or decay more rapidly. We let the current input regulate this progression coefficient:
\begin{equation}
q_t=\operatorname{SiLU}\!\left(W_q \mathbf{x}_t\right), \quad \Delta_t=\operatorname{softplus}\!\left(b_{\Delta}+W_{\Delta} q_t\right) ,
\end{equation}
where $\mathbf{x}_t \in \mathbb{R}^M$ denotes the current input, \(W_q\) and \(W_\Delta\) are learnable projection weights, and \(b_\Delta\) is a learnable bias. The neuron can therefore dynamically adjust the progression of its internal dynamics according to the current input. By the chain rule, the state dynamics with respect to external time become
\begin{equation}
\frac{\mathrm dh}{\mathrm dt}
=
\frac{\mathrm d\tau}{\mathrm dt}
\frac{\mathrm dh}{\mathrm d\tau}
=
\underbrace{\Delta_t A}_{A}h
+
\underbrace{\Delta_t B_t}_{B}x.
\end{equation}
Thus, \(\Delta_t\) directly regulates the rate at which the internal states evolve with external time.

We next specify the intrinsic dynamics of TV-MCN. The internal state \(h(t)\in\mathbb{C}^{N}\) consists of \(N\) independently evolving compartments, parameterized by a set of learnable complex-valued modes:
\begin{equation}
A=\operatorname{diag}\!\left(-e^{a_1}+j \omega_1, \ldots,-e^{a_N}+j \omega_N\right),
\end{equation}
where $a_i$ and $\omega_i$ are learnable parameters. For the $i$-th compartment, the real part $-e^{a_i}$ determines its decay rate of membrane potential, while the imaginary part $\omega_i$ determines its oscillation frequency. Different compartments therefore exhibit distinct decay scales and frequency responses. 

Beyond state evolution, $B$ and $C$ in the unified framework determine how the current input is integrated into the internal states and how these states are read out, respectively. In TV-MCN, both processes are dynamically regulated by the current input:
\begin{equation}
B_t=W_B\mathbf{x}_t,\qquad C_t^{\mathsf T}=W_C \mathbf{x}_t.
\end{equation}
The input-dependent $B_t$ allows the neuron to dynamically regulate which information is written into its internal states, while $C_t$ determines which stored information contributes to the output. 

With the internal-time modulation and input-dependent $B_t$ and $C_t$, the full form of TV-MCN is
\begin{equation}
\begin{aligned}
\underbrace{
\begin{bmatrix}
\dot h_1(t)\\
\dot h_2(t)\\
\vdots\\
\dot h_N(t)
\end{bmatrix}
}_{\dot h(t)}
&=
\Delta_t
\underbrace{
\begin{bmatrix}
-e^{a_1}+j\omega_1 & 0 & \cdots & 0\\
0 & -e^{a_2}+j\omega_2 & \cdots & 0\\
\vdots & \vdots & \ddots & \vdots\\
0 & 0 & \cdots & -e^{a_N}+j\omega_N
\end{bmatrix}
}_{A}
\underbrace{
\begin{bmatrix}
h_1(t)\\
h_2(t)\\
\vdots\\
h_N(t)
\end{bmatrix}
}_{h(t)}
+
\Delta_t
\underbrace{
\begin{bmatrix}
b_{1,t}\\
b_{2,t}\\
\vdots\\
b_{N,t}
\end{bmatrix}
}_{B_t}
x(t).
\end{aligned}
\end{equation}
\[
\begin{aligned}
u(t)
&=
\underbrace{
\begin{bmatrix}
c_{1,t}\\
c_{2,t}\\
\vdots\\
c_{N,t}
\end{bmatrix}^{\mathsf T}
}_{C_t}
\underbrace{
\begin{bmatrix}
h_1(t)\\
h_2(t)\\
\vdots\\
h_N(t)
\end{bmatrix}
}_{h(t)},\\
s(t)
&=H\!\left(u(t)-\vartheta\right),\qquad
h(t^+)=R\!\left(h(t^-),s(t)\right).
\end{aligned}
\]
where $\Delta_t$, $B_t$, and $C_t$ regulate the rate of state evolution, input integration, and state readout, respectively. For spike generation, we take the real part of the membrane potential for threshold comparison and apply a soft reset after firing.
Figure~\ref{fig:tvmcn_overview} provides an overview of these input-dependent dynamics.

To implement TV-MCN in discrete time, we adopt zero-order hold (ZOH) discretization \citep{gu2022parameterization}, assuming that the input and the dynamic parameters generated from it remain constant within each time step. Normalizing the external time-step length to $1$ gives
\begin{equation}
\bar A_t = \exp(\Delta_t A), \qquad \bar B_t = A^{-1} \left[ \exp(\Delta_t A)-I \right]B_t.
\end{equation}
Therefore, the discrete state update of TV-MCN can be written as
\begin{equation}
\begin{aligned}
h_t &= \bar A_t h_{t-1} + \bar B_t x_t,\\
u_t &= C_t h_t,\\
s_t &= H\!\left(u_t-\vartheta\right), \qquad h_{t^+}=R\!\left(h_{t^-},s_t\right).
\end{aligned}
\end{equation}
\endgroup

\section{Experimental Results}
We evaluate TV-MCN on two categories of benchmarks: temporal classification and synthetic sequence tasks. For temporal classification, we use Google Speech Commands (GSC) \citep{warden2018speech}, Spiking Heidelberg Digits (SHD) and Spiking Speech Commands (SSC) \citep{cramer2022heidelberg}, and DVS128 Gesture \citep{amir2017lowpower}, covering speech and event-based vision. For synthetic sequence tasks, we evaluate Adding \citep{hochreiter1997long}, Selective Copying \citep{gu2023mamba}, 20D STORE--RECALL \citep{salaj2021spike}, MAD ICR and MAD Noisy ICR \citep{poli2024mechanistic}, and Induction \citep{gu2023mamba}. These tasks assess accumulation, selective retention under distraction, instruction-dependent storage and retrieval, associative recall with or without irrelevant inputs, and the use of previously established temporal associations. Adding is evaluated using mean squared error (MSE), while the remaining tasks use classification or recall accuracy. Detailed experimental settings are provided in Appendix~\ref{app:experimental_settings}.

\subsection{Comparison with Existing Spiking Neurons}
We compare LIF \citep{burkitt2006review}, ad-LIF \citep{baronig2025advancing}, TC-LIF \citep{zhang2024tclif}, DH-LIF \citep{zheng2024temporal}, HetSyn \citep{deng2025hetsyn}, DGN \citep{bai2025gating}, D-RF \citep{zhang2025drf}, PMSN \citep{chen2026pmsn}, and TV-MCN using the same fully connected network architecture.

\begin{table}[!htbp]
\centering
\caption{Comparison of spiking neurons on temporal classification tasks. Accuracies are reported in \%. \textbf{Bold} and \protect\uline{underlining} indicate the best and second-best results, respectively.}
\label{tab:neuron_comparison}
\footnotesize
\setlength{\tabcolsep}{2.5pt}
\renewcommand{\arraystretch}{1.15}
\scalebox{0.9}{
\begin{tabular*}{\linewidth}{@{}>{\centering\arraybackslash}m{0.34\linewidth}|@{\extracolsep{\fill}}cccc@{}}
\toprule
Neuron & GSC & SSC & SHD & DVS128 Gesture \\
\midrule
LIF \citep{burkitt2006review} & 78.97 & 53.98 & 74.73 & 87.85 \\
ad-LIF \citep{baronig2025advancing} & 46.51 & 47.42 & 71.51 & 89.24 \\
TC-LIF \citep{zhang2024tclif} & 82.31 & 55.59 & 69.43 & 85.76 \\
DH-LIF \citep{zheng2024temporal} & 82.47 & 61.04 & 82.60 & 88.89 \\
HetSyn \citep{deng2025hetsyn} & \uline{90.00} & \uline{66.76} & \textbf{88.60} & 89.58 \\
DGN \citep{bai2025gating} & 71.73 & 63.88 & 73.85 & 84.72 \\
D-RF \citep{zhang2025drf} & 87.44 & 61.85 & 86.13 & \uline{92.01} \\
PMSN \citep{chen2026pmsn} & 87.31 & 62.16 & 84.89 & 90.62 \\
\midrule
TV-MCN & \textbf{90.56} & \textbf{70.98} & \uline{88.30} & \textbf{93.75} \\
\bottomrule
\end{tabular*}}
\vspace{4pt}

\centering
\caption{Comparison of spiking neurons on synthetic sequence tasks. Accuracies are reported in \%, except for Adding, which uses MSE. \textbf{Bold} and \protect\uline{underlining} indicate the best and second-best results, respectively.}
\label{tab:context_neuron_comparison}
\footnotesize
\setlength{\tabcolsep}{1.0pt}
\renewcommand{\arraystretch}{1.15}
\scalebox{0.9}{
\begin{tabular*}{\linewidth}{@{}>{\centering\arraybackslash}m{0.32\linewidth}|@{\hspace{3pt}\extracolsep{\fill}}cccccc@{}}
\toprule
\makecell[c]{Neuron}
& \makecell[c]{Adding\\MSE ($\times 10^{-1}$)}
& \makecell[c]{Sel. Copy.\\Acc.}
& \makecell[c]{MAD ICR\\Acc.}
& \makecell[c]{MAD Noisy\\Acc.}
& \makecell[c]{STORE--\\RECALL\\Bit}
& \makecell[c]{Induction\\Acc.} \\
\midrule
LIF \citep{burkitt2006review} & 1.6680 & 7.17 & 42.21 & 38.93 & 20.05 & 7.25 \\
ad-LIF \citep{baronig2025advancing} & 1.6670 & 7.33 & 13.92 & 12.74 & 26.03 & 6.98 \\
TC-LIF \citep{zhang2024tclif} & 1.6720 & 11.55 & 26.39 & 24.21 & 21.25 & 6.42 \\
DH-LIF \citep{zheng2024temporal} & 1.6680 & 7.09 & 30.83 & 30.89 & \uline{99.27} & \uline{21.15} \\
HetSyn \citep{deng2025hetsyn} & 1.6680 & 7.02 & \uline{97.44} & \uline{85.43} & 20.00 & 6.19 \\
DGN \citep{bai2025gating} & 1.6720 & 7.17 & 31.26 & 31.04 & 20.00 & 8.93 \\
D-RF \citep{zhang2025drf} & \uline{0.0161} & \uline{19.24} & 47.57 & 40.15 & 98.76 & 21.04 \\
PMSN \citep{chen2026pmsn} & 0.1443 & 9.58 & 39.75 & 34.77 & 90.13 & 20.26 \\
\midrule
TV-MCN & \textbf{0.0022} & \textbf{93.46} & \textbf{100.00} & \textbf{99.99} & \textbf{99.92} & \textbf{97.71} \\
\bottomrule
\end{tabular*}}
\end{table}

Tables~\ref{tab:neuron_comparison} and~\ref{tab:context_neuron_comparison} compare TV-MCN with existing spiking neurons on temporal classification and synthetic sequence tasks, respectively. On temporal classification, TV-MCN achieves the best accuracy on GSC, SSC, and DVS128 Gesture, reaching 90.56\%, 70.98\%, and 93.75\%, respectively, while achieving 88.30\% on SHD, close to the best-performing HetSyn. The competitive performance of existing neurons suggests that conventional temporal classification can already be handled effectively by rich neuronal dynamics, while TV-MCN further improves performance on most benchmarks. A much larger advantage emerges on synthetic sequence tasks that require selective integration, maintenance, and use of information. TV-MCN achieves the best performance on all six tasks, with particularly large gains on Selective Copying, MAD ICR, MAD Noisy ICR, and Induction. Although DH-LIF also performs strongly on STORE--RECALL, reaching 99.27\% compared with 99.92\% for TV-MCN, this strong performance does not extend to the other synthetic tasks. More generally, existing neurons tend to perform well on individual tasks but do not maintain consistently strong performance across different information-processing requirements. In contrast, TV-MCN performs consistently well across all six benchmarks, demonstrating stronger capability in handling different forms of temporal information processing.

\begin{figure}[!htbp]
    \centering
    \includegraphics[width=\linewidth]{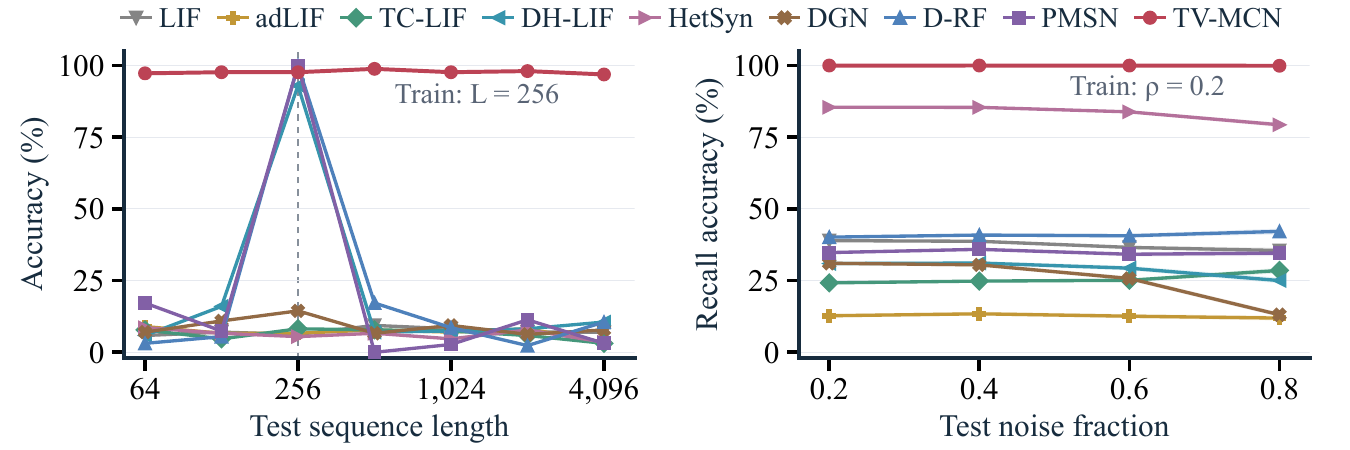}
    \caption{Generalization to sequence length and interference. Left: Length generalization on Induction. The same trained model is evaluated at different sequence lengths, with the dashed line indicating the training length. Right: Generalization across noise fractions $\rho$ on MAD Noisy ICR. Models trained at $\rho=0.2$ are evaluated at all four noise fractions without retraining.}
\label{fig:induction_noise}
\end{figure}

\begin{figure}[!htbp]
\centering
\includegraphics[width=0.85\linewidth]{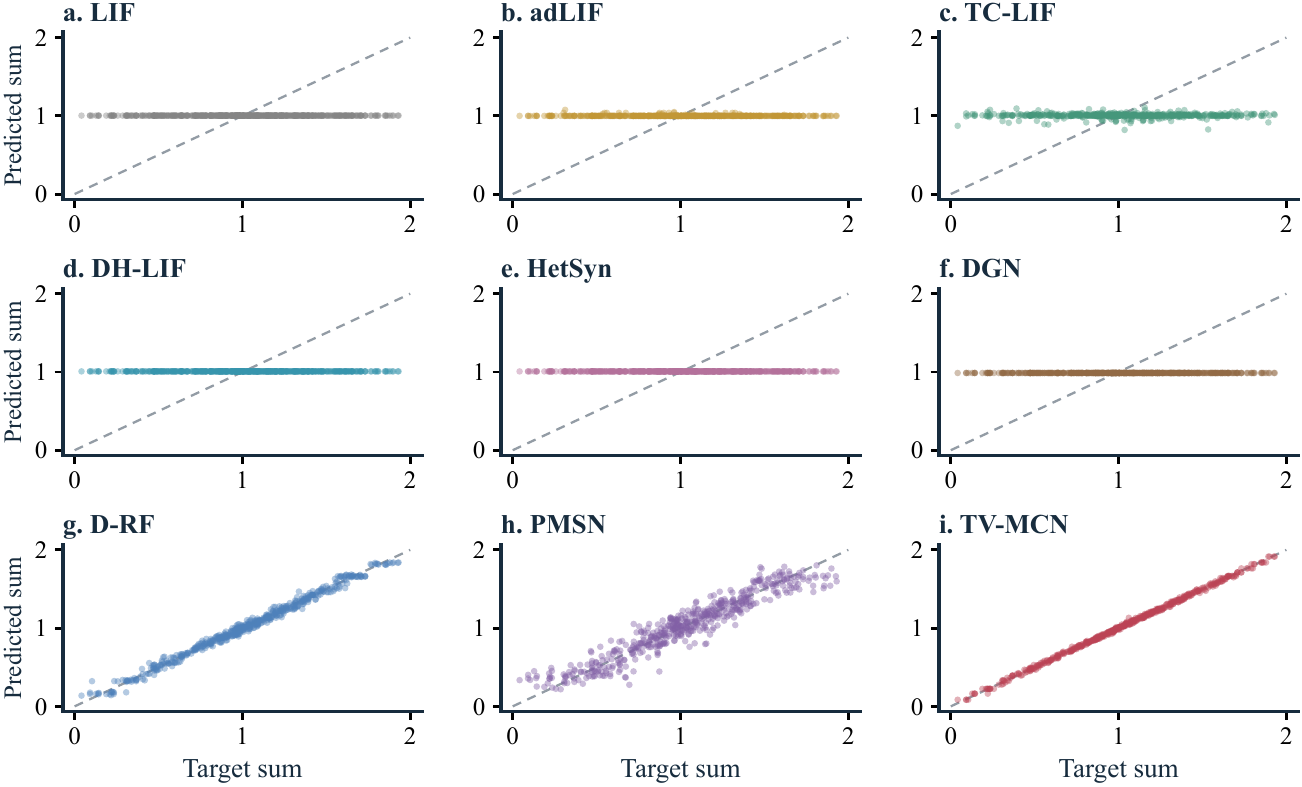}
\caption{Adding predictions for nine neuron models under the shared two-layer fully connected network.}
\label{fig:adding_predictions}
\end{figure}

We next vary sequence length and interference strength to examine whether the learned temporal processing behavior generalizes when information must be retained for different durations or withstand stronger interference. On Induction, we train each model at a fixed sequence length and evaluate them on sequences of different lengths. This setting tests whether the ability to retain and retrieve relevant historical information according to the current input generalizes beyond the training length. As shown in Figure~\ref{fig:induction_noise} (Left), PMSN and D-RF perform accurately at the training length but deteriorate substantially on shorter or longer sequences. TV-MCN maintains 96.88--98.83\% accuracy throughout the tested range, demonstrating substantially stronger length generalization. On MAD Noisy ICR, we vary the interference fraction $\rho$ to evaluate robustness to increasing interference. Each model is trained at $\rho=0.2$ and evaluated at $\rho\in\{0.2,0.4,0.6,0.8\}$ without further training. This setting tests whether relevant information can still be retained and retrieved as the amount of irrelevant input increases beyond the training condition. As shown in Figure~\ref{fig:induction_noise} (Right), TV-MCN achieves 99.99\%, 99.99\%, 99.97\%, and 99.90\% token accuracy, respectively. At $\rho=0.8$, the strongest baseline, HetSyn, reaches 79.33\%. These results indicate that TV-MCN maintains high recall accuracy under substantially stronger interference.

We also examine whether neurons can selectively accumulate input values and maintain the result until the final output. In Adding, the model must sum the marked values while ignoring the remaining inputs. As shown in Figure~\ref{fig:adding_predictions}, TV-MCN's predictions follow the target sums more closely than those of the other neurons, as indicated by their closer alignment with $y=x$. This observation is consistent with its lower MSE in Table~\ref{tab:context_neuron_comparison}. These results suggest that TV-MCN's advantage extends beyond sequence recall to maintaining selected numerical information and using it to compute the final sum despite intervening irrelevant inputs.

\FloatBarrier
\subsection{Comparison with State-of-the-Art Methods}
\begin{table}[!htbp]
\centering
\caption{Comparison with state-of-the-art methods on synthetic sequence tasks. \textbf{Bold} and \protect\uline{underlining} indicate the best and second-best results, respectively.}
\label{tab:sota_comparison}
\footnotesize
\setlength{\tabcolsep}{2.5pt}
\renewcommand{\arraystretch}{1}
\scalebox{0.9}{
\begin{tabular*}{\linewidth}{@{\extracolsep{\fill}}lcccc@{}}
\toprule
\makecell[l]{Task} & \makecell[c]{Model} & \makecell[c]{Params\\(M)} & \makecell[c]{Est. FLOPs\\(G/seq)} & \makecell[c]{Performance} \\
\midrule
\multirow{6}{*}{\shortstack[l]{MAD ICR\\Acc. (\%)}}
& PMSN \citep{chen2026pmsn} & \uline{0.291} & \uline{0.098} & 73.97 \\
& D-RF \citep{zhang2025drf} & 3.298 & 1.059 & \uline{99.87} \\
& SpikingSSMs \citep{shen2025spikingssms} & 35.804 & 10.491 & 98.98 \\
& S5-RF \citep{huber2024scaling} & 1.612 & 0.887 & 39.18 \\
& Spiking LMUFormer \citep{liu2024lmuformer} & 0.802 & 0.207 & 57.65 \\
\cmidrule(lr){2-5}
& TV-MCN & \textbf{0.062} & \textbf{0.026} & \textbf{100.00} \\
\midrule
\multirow{6}{*}{\shortstack[l]{MAD Noisy ICR\\Acc. (\%)}}
& PMSN \citep{chen2026pmsn} & \uline{0.294} & \uline{0.098} & 68.16 \\
& D-RF \citep{zhang2025drf} & 3.331 & 1.068 & \uline{99.85} \\
& SpikingSSMs \citep{shen2025spikingssms} & 35.836 & 10.495 & 96.96 \\
& S5-RF \citep{huber2024scaling} & 1.645 & 0.904 & 30.78 \\
& Spiking LMUFormer \citep{liu2024lmuformer} & 0.811 & 0.208 & 50.37 \\
\cmidrule(lr){2-5}
& TV-MCN & \textbf{0.066} & \textbf{0.026} & \textbf{99.99} \\
\midrule
\multirow{6}{*}{\shortstack[l]{Selective Copying\\Acc. (\%)}}
& PMSN \citep{chen2026pmsn} & \uline{0.291} & \uline{3.500} & 19.10 \\
& D-RF \citep{zhang2025drf} & 3.298 & 35.244 & 18.68 \\
& SpikingSSMs \citep{shen2025spikingssms} & 35.804 & 342.255 & \textbf{97.83} \\
& S5-RF \citep{huber2024scaling} & 1.612 & 29.864 & 46.68 \\
& Spiking LMUFormer \citep{liu2024lmuformer} & 0.802 & 6.723 & 7.26 \\
\cmidrule(lr){2-5}
& TV-MCN & \textbf{0.062} & \textbf{0.827} & \uline{95.41} \\
\midrule
\multirow{6}{*}{\shortstack[l]{20D STORE--RECALL\\Bit Acc. (\%)}}
& PMSN \citep{chen2026pmsn} & \uline{0.300} & \uline{1.724} & 97.83 \\
& D-RF \citep{zhang2025drf} & 3.409 & 17.547 & \textbf{100.00} \\
& SpikingSSMs \citep{shen2025spikingssms} & 35.880 & 166.997 & \uline{99.99} \\
& S5-RF \citep{huber2024scaling} & 1.690 & 15.084 & 98.04 \\
& Spiking LMUFormer \citep{liu2024lmuformer} & 0.822 & 3.377 & 20.00 \\
\cmidrule(lr){2-5}
& TV-MCN & \textbf{0.072} & \textbf{0.457} & \uline{99.99} \\
\midrule
\multirow{6}{*}{\shortstack[l]{Induction\\Acc. (\%)}}
& PMSN \citep{chen2026pmsn} & \uline{0.291} & \uline{0.972} & 22.43 \\
& D-RF \citep{zhang2025drf} & 3.298 & 9.924 & 17.47 \\
& SpikingSSMs \citep{shen2025spikingssms} & 35.804 & 96.697 & 21.54 \\
& S5-RF \citep{huber2024scaling} & 1.612 & 8.391 & \uline{59.10} \\
& Spiking LMUFormer \citep{liu2024lmuformer} & 0.802 & 1.901 & 6.92 \\
\cmidrule(lr){2-5}
& TV-MCN & \textbf{0.062} & \textbf{0.233} & \textbf{100.00} \\
\midrule
\multirow{6}{*}{\shortstack[l]{Adding\\MSE ($\times 10^{-2}$)}}
& PMSN \citep{chen2026pmsn} & \uline{0.288} & \uline{0.818} & 0.1210 \\
& D-RF \citep{zhang2025drf} & 3.268 & 8.434 & \uline{0.0202} \\
& SpikingSSMs \citep{shen2025spikingssms} & 35.773 & 83.053 & 0.0275 \\
& S5-RF \citep{huber2024scaling} & 1.582 & 7.044 & 0.5470 \\
& Spiking LMUFormer \citep{liu2024lmuformer} & 0.795 & 1.630 & 8.1600 \\
\cmidrule(lr){2-5}
& TV-MCN & \textbf{0.058} & \textbf{0.196} & \textbf{0.0181} \\
\bottomrule
\end{tabular*}}
\end{table}

In this section, we compare TV-MCN with representative full architectures of recent methods and report both parameter counts and computational costs to examine whether its performance gains can be attributed simply to increased model size or computational complexity. The compared methods include PMSN \citep{chen2026pmsn}, D-RF \citep{zhang2025drf}, SpikingSSMs \citep{shen2025spikingssms}, S5-RF \citep{huber2024scaling}, and Spiking LMUFormer \citep{liu2024lmuformer}. TV-MCN adopts a compact two-layer convolution-augmented spiking sequence network, with architectural details provided in Appendix~\ref{app:experimental_settings}. For each method, the core network architecture is kept unchanged across tasks, with only the input and output layers adapted to the task.

Table~\ref{tab:sota_comparison} reports performance, parameter counts, and FLOPs per sequence on the six synthetic sequence tasks. Details of the calculation procedure are provided in Appendix~\ref{app:computational_cost}. TV-MCN achieves strong performance across the synthetic sequence tasks while using the fewest parameters and the lowest FLOPs among the compared methods. For example, on Induction, TV-MCN achieves 100.00\% accuracy with 0.062~M parameters and 0.233~G estimated FLOPs per sequence, compared with 17.47\% for D-RF and 59.10\% for S5-RF, both of which require substantially more parameters and computation. These results demonstrate a favorable performance--cost trade-off for TV-MCN across the evaluated tasks. In particular, its performance gains cannot be simply attributed to a larger model or greater computation.

\FloatBarrier
\subsection{Ablation Studies}
To examine the role of time-variant regulation in selective retention and retrieval, we separately remove the two modulation mechanisms of TV-MCN on Selective Copying while retaining its multiscale state structure. Fixing $\Delta_t=1$ removes input-dependent control of the state-evolution rate, while replacing $B_t,C_t$ with learnable static parameters $B_0,C_0$ removes dynamic regulation of input integration and state readout. The network architecture, state size, and training budget remain unchanged.

\begin{table}[!htbp]
\centering
\caption{Component ablation on Selective Copying. Exact accuracy requires all target tokens to be correct. \textbf{Bold} indicates the best results.}
\label{tab:ablation}
\footnotesize
\setlength{\tabcolsep}{7pt}
\renewcommand{\arraystretch}{0.95}
\begin{tabular}{ccc}
\toprule
Model Variant & Token Acc. (\%) & Exact Acc. (\%) \\
\midrule
\textbf{TV-MCN} & \textbf{93.46} & \textbf{37.72} \\
w/o internal time modulation & 7.20 & 0.00 \\
w/o dynamic $B/C$ coupling & 61.24 & 0.02 \\
\bottomrule
\end{tabular}
\end{table}

Table~\ref{tab:ablation} shows that removing internal-time modulation reduces token accuracy from 93.46\% to 7.20\%, while using static $B/C$ reduces it to 61.24\%. Exact accuracy drops from 37.72\% to 0.00\% and 0.02\%, respectively. Static $B/C$ preserves partial token-level recall but rarely recovers a complete target sequence. Both variants retain the multiscale states and the other modulation mechanism, yet suffer substantial recall losses.

\begin{figure}[!htbp]
\centering
\includegraphics[width=0.96\linewidth]{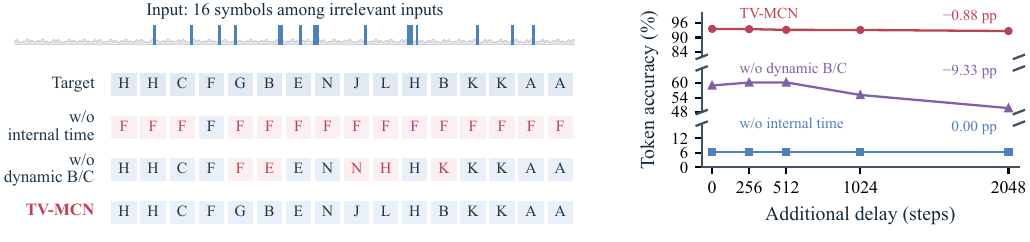}
\caption{Recall behavior after component removal. Left: Target and outputs of the three variants for the same input; red indicates errors. Right: Token accuracy after inserting additional irrelevant inputs before recall for the same samples.}
\label{fig:sc_ablation_recall}
\end{figure}

Figure~\ref{fig:sc_ablation_recall} (left) illustrates these effects for the same input: without internal-time modulation, the output repeats a single token; with static $B/C$, several target tokens are recalled incorrectly. We then insert additional irrelevant inputs before recall to examine how these removals affect retention under continued interference. The three trained models are evaluated on the same samples without retraining, with targets unchanged. With 2048 additional inputs, token accuracy drops by 0.88 percentage points for the full model and 9.33 for static $B/C$, while the variant without internal-time modulation remains at 6.45\% (right). The larger decline with static $B/C$, despite retaining internal-time modulation, supports the contribution of dynamic integration and readout to maintaining recall accuracy under interference.

\begin{figure}[!htbp]
\centering
\captionsetup{font=small}
\includegraphics[width=0.85\linewidth]{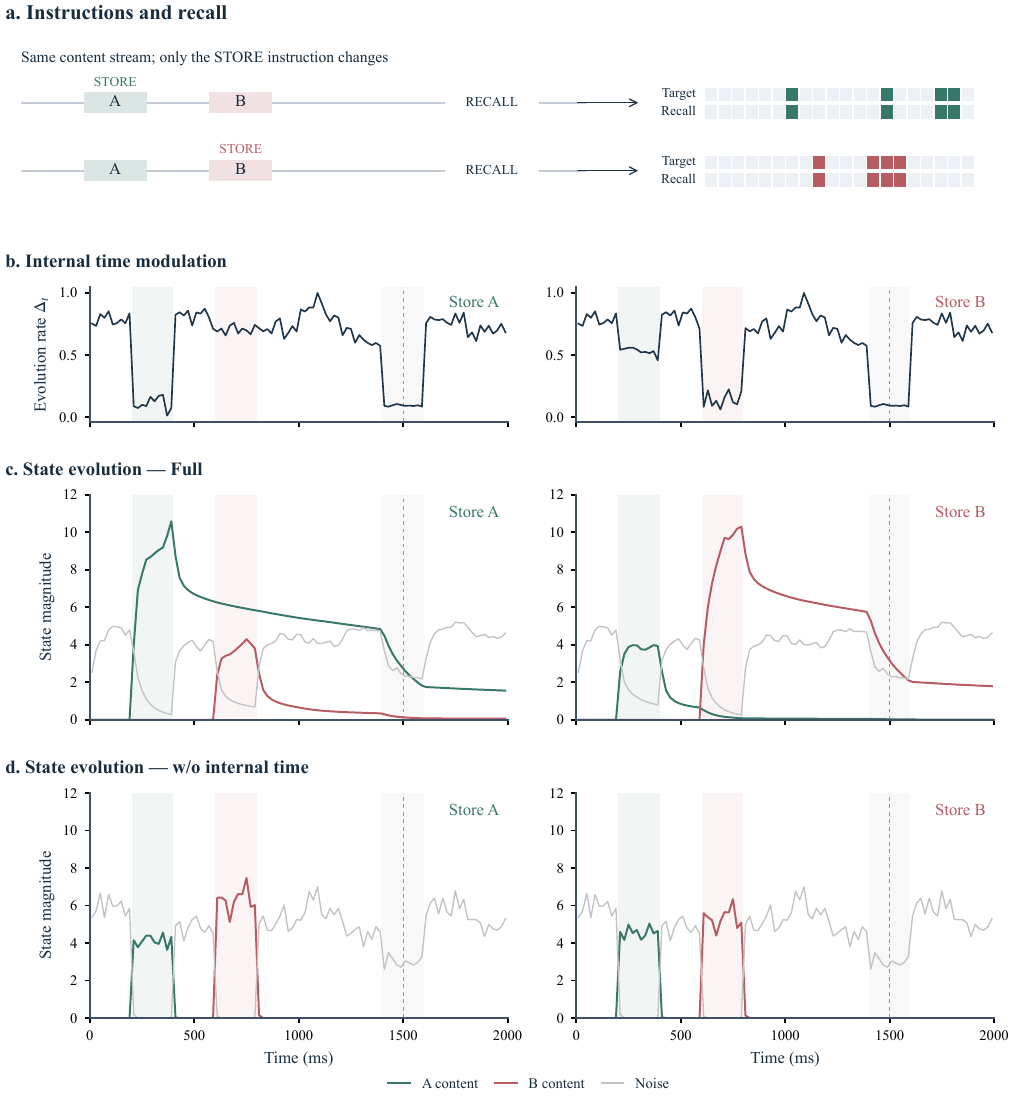}
\caption{Effect of internal-time modulation on instruction-dependent retention. (a) The input content is identical; only the STORE instruction changes, requiring the model to remember either A or B and recall it later. (b) The input-dependent rate of internal-state evolution under the two instructions. (c--d) The full TV-MCN and the model without internal-time modulation are compared to show how strongly A, B, and irrelevant content are retained in the internal states. The left and right columns correspond to remembering A and B, respectively.}
\label{fig:store_state_dynamics}
\end{figure}

We further examine the role of internal-time modulation in maintaining information selected by an instruction. On STORE--RECALL, we compare the full model with the variant without internal-time modulation; both retain dynamic $B/C$. The content sequence is identical, while the STORE instruction selects either A or B for later recall (Figure~\ref{fig:store_state_dynamics}(a)).

Figure~\ref{fig:store_state_dynamics}(b) shows that the full model lowers $\Delta_t$ during the instructed STORE period, slowing state evolution at that time. In the full model, the state contribution of the selected content persists until recall, while that of the unselected content decays more rapidly (c). Without internal-time modulation, both contributions decay rapidly after their respective inputs end, regardless of which content is selected (d). These differences support the role of internal-time modulation in maintaining selected content.

Together, these ablations show that removing either internal-time modulation or dynamic $B/C$ impairs recall despite retaining the multiscale state structure. These results support the contribution of time-variant regulation to the selective integration, maintenance, and use of information.

\section{Conclusion}

We introduced a biologically inspired unified state-dynamics framework for analyzing and comparing spiking neuron designs. The framework reveals that although existing designs enrich neuronal dynamics, the rules determining input integration, state evolution, and state readout generally remain unchanged during temporal processing. To address this limitation, TV-MCN introduces time-variant regulation into multiscale compartment dynamics: input-dependent internal time regulates state evolution, while dynamic $B_t$ and $C_t$ regulate input integration and state readout. Experiments show competitive performance on temporal classification and substantial gains on synthetic sequence tasks requiring selective retention and retrieval. Generalization tests and ablations further support these time-varying mechanisms. Together, the results highlight time-variant neuronal dynamics as an effective approach for adjusting how information is integrated, maintained, and used over time.

\bibliography{references}
\bibliographystyle{references}

\appendix
\section{Complete Framework Formulations of Representative Neurons}
\label{app:framework_formulations}
The internal dynamics of spiking neurons can be summarized by the following framework from a state-space perspective:
\[
\begin{aligned}
\dot{h}(t) & =A h(t)+B x(t) \\
u(t) & =C h(t) \\
s(t) & =H\left(u\left(t\right)-\vartheta(t)\right),\qquad h\left(t^{+}\right) =R\left(h\left(t^{-}\right), s(t)\right).
\end{aligned}
\]
We group neuron models according to whether the state-evolution term $Ah$ is linear or nonlinear in the internal state $h$. In the models considered here, this distinction is reflected in whether $A$ depends on $h$. Here, $x$ denotes the input as defined in each model's framework formulation. For each model, we first present its original dynamics, then express them in the unified framework and analyze how its state structure and dynamical rules affect temporal processing.

\subsection{Neuron Models with Linear State Evolution}
\subsubsection*{1. LIF \citep{burkitt2006review}}
For a LIF neuron with zero resting potential, the continuous-time form is
\[
\tau_m \dot{u}(t)=-u(t)+W x(t),
\]
where $x(t)$ is the input and $W$ is the input weight of the neuron. Dividing both sides by $\tau_m$ gives
\[
\dot{u}(t)=-\frac{1}{\tau_m} u(t)+\frac{1}{\tau_m} W x(t) .
\]
The membrane potential $u(t)$ is the only internal state in this LIF model. We therefore define
\[
h(t)=u(t) .
\]
Therefore, LIF can be written as
\[
\begin{aligned}
\underbrace{\dot{u}(t)}_{\dot{h}} & =\underbrace{-\frac{1}{\tau_m}}_A \underbrace{u(t)}_h+\underbrace{\frac{1}{\tau_m} W}_B x(t), \\
u(t) & =\underbrace{1}_C \underbrace{u(t)}_h, \\
s(t) & =H(u(t)-\vartheta),\qquad u\left(t^{+}\right) =R\left(h\left(t^{-}\right), s(t)\right) .
\end{aligned}
\]
Thus, LIF corresponds to a one-dimensional state system: $A=-1/\tau_m$ specifies exponential decay of the membrane potential with time constant $\tau_m$; $B=W/\tau_m$ represents input integration and contains the trainable synaptic weight $W$; $C=1$ reads out the membrane potential; and $R$ describes soft or hard reset after a spike. In this LIF model, the decay time constant $\tau_m$ is preset and not learned during training. The decay and input-integration coefficients remain unchanged throughout the sequence.
\subsubsection*{2. Spike Response Model (SRM) \citep{gerstner1995time,gerstner2014neuronal}}
The SRM with exponential response kernels uses a separate trace for each input spike train and another trace for the neuron's own firing history. The trace for input $j$ is
\[
q_j(t)=\sum_{t_j^f<t}e^{-(t-t_j^f)/\tau_j}.
\]
Here, $t_j^f$ denotes the time of the $f$-th spike from input $j$. Each spike adds a unit-amplitude contribution to $q_j$, which then decays exponentially with time constant $\tau_j$. Therefore, between spikes, $q_j$ decays with the same time constant, while each new input spike increases it by $1$. This gives the following state update:
\[
\dot q_j=-\frac{q_j}{\tau_j}+x_j(t),
\]
where $x_j(t)=\sum_f\delta(t-t_j^f)$ is the spike train of input $j$.

The neuron's own firing history is retained in another trace $r$:
\[
r(t)=\sum_{t^f<t}e^{-(t-t^f)/\tau_r}.
\]
Here, $t^f$ denotes the time of a spike generated by the neuron. Like the input traces, $r$ decays between spikes and increases by $1$ whenever the neuron fires:
\[
\dot r=-\frac{r}{\tau_r},\qquad r(t^+)=r(t^-)+1\quad\text{(at a spike)}.
\]

The membrane potential is determined jointly by these two types of historical responses. With zero resting potential, the SRM membrane potential is
\[
\begin{aligned}
u(t)&=\sum_{j=1}^{M}w_j\sum_{t_j^f<t}\epsilon_j(t-t_j^f)
+\sum_{t^f<t}\eta(t-t^f)\\
&=\sum_{j=1}^{M}w_jq_j(t)-\kappa r(t).
\end{aligned}
\]
Here, $w_j$ is the synaptic weight, and $\xi\geq0$ denotes the time elapsed since a spike. The response to an input spike is $\epsilon_j(\xi)=e^{-\xi/\tau_j}$. The inhibitory response following the neuron's own spike is $\eta(\xi)=-\kappa e^{-\xi/\tau_r}$, where $\kappa>0$ controls the strength of inhibition.

The membrane potential is thus determined directly by these traces, and the internal states evolving over time are the input traces $q_j$ and the firing trace $r$:
\[
h(t)=\begin{bmatrix}q_1(t)\\\vdots\\q_M(t)\\r(t)\end{bmatrix},\qquad
x(t)=\begin{bmatrix}x_1(t)\\\vdots\\x_M(t)\end{bmatrix}.
\]
The SRM with exponential kernels can therefore be written in the following framework form:
\begingroup
\allowdisplaybreaks[1]
\begin{align*}
\underbrace{\begin{bmatrix}\dot q_1\\\vdots\\\dot q_M\\\dot r\end{bmatrix}}_{\dot h(t)}
&=
\underbrace{
\begin{bmatrix}
-\tau_1^{-1}&0&\cdots&0\\
0&\ddots&\ddots&\vdots\\
\vdots&\ddots&-\tau_M^{-1}&0\\
0&\cdots&0&-\tau_r^{-1}
\end{bmatrix}}_A
\underbrace{\begin{bmatrix}q_1\\\vdots\\q_M\\r\end{bmatrix}}_{h(t)}
+\underbrace{\begin{bmatrix}I_M\\\mathbf 0^{\mathsf T}\end{bmatrix}}_B
\underbrace{\begin{bmatrix}x_1\\\vdots\\x_M\end{bmatrix}}_{x(t)},\\[10pt]
u(t)&=
\underbrace{\begin{bmatrix}w_1&\cdots&w_M&-\kappa\end{bmatrix}}_C
\underbrace{\begin{bmatrix}q_1\\\vdots\\q_M\\r\end{bmatrix}}_{h(t)},\\[10pt]
s(t)&=H\!\left(u(t)-\vartheta\right),\qquad
h(t^+)=R\!\left(h(t^-),s(t)\right)
=\begin{bmatrix}q_1^-\\\vdots\\q_M^-\\r^-+s(t)\end{bmatrix}.
\end{align*}
\endgroup
From a state-space perspective, the SRM with fixed exponential kernels corresponds to a set of independent real-valued decay states. Choosing the response kernels specifies how these states evolve over time, as reflected in the diagonal entries of $A$. Input spikes update the corresponding states through $B$, and $C$ weights these states to form the membrane potential. The trace values evolve with incoming and self-generated spikes, while their decay time constants and readout weights are time-invariant.

\subsubsection*{3. ad-LIF \citep{baronig2025advancing}}
The continuous-time form of ad-LIF is
\[
\begin{aligned}
\tau_u \dot{u}(t) & =-u(t)+I(t)-w(t), \\
\tau_w \dot{w}(t) & =-w(t)+a u(t)+b z(t).
\end{aligned}
\]
In addition to the membrane potential $u$, ad-LIF introduces an adaptation state $w$ that must be retained across time. The internal state therefore expands from one to two dimensions:
\[
h= \begin{bmatrix} u\\ w \end{bmatrix}.
\]
In addition to the external input current $I(t)$, the neuron's own spike train $z(t)$ drives the adaptation state. The driving input to the state dynamics therefore becomes two-dimensional:
\[
x = \begin{bmatrix} I\\ z \end{bmatrix}.
\]
Thus, ad-LIF can be written in the following form:
\[
\begin{aligned}
\underbrace{
\begin{bmatrix}
\dot{u}\\
\dot{w}
\end{bmatrix}}_{\dot{h}}
&=
\underbrace{
\begin{bmatrix}
-\dfrac{1}{\tau_u} & -\dfrac{1}{\tau_u}\\
\dfrac{a}{\tau_w} & -\dfrac{1}{\tau_w}
\end{bmatrix}}_{A}
\underbrace{
\begin{bmatrix}
u\\
w
\end{bmatrix}}_{h}
+
\underbrace{
\begin{bmatrix}
\dfrac{1}{\tau_u} & 0\\
0 & \dfrac{b}{\tau_w}
\end{bmatrix}}_{B}
\underbrace{
\begin{bmatrix}
I\\
z
\end{bmatrix}}_{x},
\\[8pt]
u(t)
&=
\underbrace{
\begin{bmatrix}
1 \\
0
\end{bmatrix}^{\mathsf T}}_{C}
\underbrace{
\begin{bmatrix}
u\\
w
\end{bmatrix}}_{h},
\\[8pt]
s(t) &=H\bigl(u(t)-\vartheta\bigr),\qquad h(t^{+}) =R\bigl(h(t^{-}),s(t)\bigr).
\end{aligned}
\]
In implementations with learnable time constants, $\tau_u$ and $\tau_w$ are optimized during training. They therefore determine learnable coefficients in $A$ and the corresponding entries of $B$, while $C=[1,0]$ remains a fixed readout of the membrane potential.

From a state-space perspective, ad-LIF extends the single membrane-potential state of LIF to two coupled states, the membrane potential $u$ and the adaptation state $w$. The membrane potential drives the accumulation of $w$, while $w$ suppresses $u$ through negative feedback, introducing an additional slow time scale and history-dependent adaptation. Compared with the single decay mode of LIF, the two-dimensional coupled matrix $A$ therefore supports richer temporal dynamics. However, once learned, the coefficients governing these dynamics remain unchanged during temporal processing.
\subsubsection*{4. TC-LIF \citep{zhang2024tclif}}
TC-LIF is defined directly in discrete time. It contains a dendritic compartment $U^D$ and a somatic compartment $U^S$, with the following updates:
\[
\begin{aligned}
U^D[t] & =U^D[t-1]+\beta_1 U^S[t-1]+I[t]-\gamma S[t-1], \\
U^S[t] & =U^S[t-1]+\beta_2 U^D[t]-V_{\mathrm{th}} S[t-1], \\
S[t] & =\Theta\left(U^S[t]-V_{\mathrm{th}}\right),
\end{aligned}
\]
where
\[
\beta_1=-\sigma\left(c_1\right) \in(-1,0), \quad \beta_2=\sigma\left(c_2\right) \in(0,1) .
\]
Compared with LIF, TC-LIF adds a dendritic state that also evolves over time, expanding the internal state from one to two dimensions:
\[
h[t]=\left[\begin{array}{l}
U^D[t] \\
U^S[t]
\end{array}\right] .
\]
In addition to the external input current $I[t]$, the spike $S[t-1]$ generated at the previous time step also drives the updates of both internal states. The driving input to the state dynamics therefore becomes two-dimensional:
\[
x[t]=\left[\begin{array}{c}
I[t] \\
S[t-1]
\end{array}\right] .
\]
Following the update order in the original formulation, \(U^D[t]\) is first updated and the resulting value is then used to update \(U^S[t]\). Substituting the former update into the latter gives
\[
U^S[t]=\beta_2 U^D[t-1]+\left(1+\beta_1 \beta_2\right) U^S[t-1]+\beta_2 I[t]-\left(\beta_2 \gamma+V_{\mathrm{th}}\right) S[t-1].
\]
Therefore, TC-LIF can be written in the following unified form:
\[
\begin{aligned}
\underbrace{\left[\begin{array}{l}
U^D[t] \\
U^S[t]
\end{array}\right]}_{h[t]} & =\underbrace{\left[\begin{array}{cc}
1 & \beta_1 \\
\beta_2 & 1+\beta_1 \beta_2
\end{array}\right]}_A \underbrace{\left[\begin{array}{l}
U^D[t-1] \\
U^S[t-1]
\end{array}\right]}_{h[t-1]}+\underbrace{\left[\begin{array}{cc}
1 & -\gamma \\
\beta_2 & -\left(\beta_2 \gamma+V_{\mathrm{th}}\right)
\end{array}\right]}_B \underbrace{\left[\begin{array}{c}
I[t] \\
S[t-1]
\end{array}\right]}_{x[t]}, \\
u[t] & =\underbrace{\left[\begin{array}{c}
0 \\
1
\end{array}\right]^{\mathsf T}}_C \underbrace{\left[\begin{array}{l}
U^D[t] \\
U^S[t]
\end{array}\right]}_{h[t]}, \\
s[t] & =H\left(u[t]-V_{\mathrm{th}}\right),\qquad h\left[t^{+}\right] =R\left(h\left[t^{-}\right], s[t]\right)=h\left[t^{-}\right] .
\end{aligned}
\]
In TC-LIF, $\beta_1=-\sigma(c_1)$ and $\beta_2=\sigma(c_2)$ are parameterized by the learnable parameters $c_1$ and $c_2$. Therefore, $A$ and the entries of $B$ that depend on $\beta_2$ are adjusted during training, while $C=[0,1]$ remains a fixed readout of the somatic state. The recovery terms driven by the previous spike are included in the second column of $B$, so no additional post-spike state update is required in this representation.

From a state-space perspective, TC-LIF extends the one-dimensional state of LIF to two coupled states and, more importantly, removes their independent leakage while constraining $\beta_1<0$ and $\beta_2>0$. These constraints give $\det A=1$ and a pair of complex-conjugate eigenvalues with unit magnitude. The resulting state modes therefore oscillate without exponential decay, rather than exhibiting the damped dynamics of LIF or general leaky two-state systems. Correspondingly, the eigenvalue magnitudes of $A^k$ do not decay exponentially with $k$ during backpropagation through time, helping alleviate vanishing gradients over long sequences. Despite these richer dynamics, once the parameters are learned, the coefficients governing state evolution, input integration, and state readout remain unchanged during temporal processing.
\subsubsection*{5. DH-LIF \citep{zheng2024temporal}}
In addition to the somatic membrane potential of LIF, DH-LIF introduces a dendritic current for each dendritic branch that can preserve information across time. The paper uses the following discrete form:
\[
\begin{aligned}
u[t+1] & =\beta u[t]+(1-\beta) \sum_{d=1}^D i_d[t+1]-u_{\mathrm{th}} o[t] \\
i_d[t+1] & =\alpha_d i_d[t]+\left(1-\alpha_d\right) I_d[t+1] \\
o[t+1] & =H\left(u[t+1]-u_{\mathrm{th}}\right).
\end{aligned}
\]
Compared with LIF, DH-LIF adds $D$ dendritic currents that persist across time, expanding the internal state from one to $D+1$ dimensions:
\[
h[t]=\left[\begin{array}{c}
i_1[t] \\
\vdots \\
i_D[t] \\
u[t]
\end{array}\right] .
\]
Different dendritic branches receive their respective synaptic inputs. With the spike-driven reset handled separately by $R$, the state input is
\[
x[t+1]=\left[\begin{array}{c}
I_1[t+1] \\
\vdots \\
I_D[t+1]
\end{array}\right] .
\]
Substituting the dendritic update into the soma update and separating the soft-reset term gives the intermediate state $\tilde h[t+1]$:
\begingroup
\setlength{\arraycolsep}{2pt}
\[
\resizebox{\linewidth}{!}{$\displaystyle
\begin{aligned}
\underbrace{
\begin{bmatrix}
i_1[t+1] \\
\vdots \\
i_D[t+1] \\
\tilde u[t+1]
\end{bmatrix}
}_{\tilde h[t+1]}
&=
\underbrace{
\begin{bmatrix}
\alpha_1 & 0 & \cdots & 0 \\
0 & \ddots & \ddots & \vdots \\
\vdots & \ddots & \alpha_D & 0 \\
(1-\beta)\alpha_1 & \cdots & (1-\beta)\alpha_D & \beta
\end{bmatrix}
}_{A}
\underbrace{
\begin{bmatrix}
i_1[t] \\
\vdots \\
i_D[t] \\
u[t]
\end{bmatrix}
}_{h_{[t]}}
+
\underbrace{
\begin{bmatrix}
1-\alpha_1 & \cdots & 0 \\
\vdots & \ddots & \vdots \\
0 & \cdots & 1-\alpha_D \\
(1-\beta)(1-\alpha_1) & \cdots & (1-\beta)(1-\alpha_D)
\end{bmatrix}
}_{B}
\underbrace{
\begin{bmatrix}
I_1[t+1] \\
\vdots \\
I_D[t+1]
\end{bmatrix}
}_{x[t+1]},
\end{aligned}
$}
\]
\endgroup
\[
\begin{aligned}
u[t+1]
&=
\underbrace{
\begin{bmatrix}
0 \\
\vdots \\
0 \\
1
\end{bmatrix}^{\mathsf T}
}_{C}
\underbrace{
\begin{bmatrix}
i_1[t+1] \\
\vdots \\
i_D[t+1] \\
u[t+1]
\end{bmatrix}
}_{h_{[t+1]}},
\\[10pt]
\underbrace{o[t+1]}_{s[t+1]} &= H\bigl(u[t+1]-u_{\mathrm{th}}\bigr),\qquad h[t+1]=R\!\left(\tilde h[t+1],s[t]\right).
\end{aligned}
\]
Here, $s[t]=o[t]$ and $R(\tilde h,s)=\tilde h-u_{\mathrm{th}}e_{D+1}s$, with $e_{D+1}=[0,\ldots,0,1]^{\mathsf T}$ selecting the somatic state. 

In DH-LIF, the dendritic and somatic time constants are learnable, so $A$ and the entries of $B$ determined by $\alpha_d$ and $\beta$ are adjusted during training. The readout $C=[0,\ldots,0,1]$ always selects the somatic membrane potential.

From a state-space perspective, DH-LIF extends LIF by introducing multiple dendritic current states with different decay coefficients. In $A$, each $\alpha_d$ specifies the real decay mode of one dendritic branch, while $\beta$ determines the decay of the somatic state. Different $\alpha_d$ therefore provide different memory time scales. Compared with ad-LIF and TC-LIF, which enrich temporal dynamics mainly through coupling between a small number of states, DH-LIF expands the temporal basis by introducing multiple heterogeneous decay modes. Despite this richer state structure, the coefficients governing state evolution, input integration, and state readout remain unchanged during temporal processing.
\subsubsection*{6. HetSyn \citep{deng2025hetsyn}}
HetSyn treats the synaptic current of each input synapse as an independent state. Its continuous-time form is
\[
\begin{aligned}
V(t) & =\sum_{i=1}^M I_i(t)-J(t), \\
\frac{\mathrm{d} I_i(t)}{\mathrm{d} t} & =-\frac{1}{\tau_{s, i}} I_i(t)+w_i \sum_j \delta\left(t-t_{j i}\right), \\
\frac{\mathrm{d} J(t)}{\mathrm{d} t} & =-\frac{1}{\tau_J} J(t)+\vartheta \sum_j \delta\left(t-t_{j s}\right).
\end{aligned}
\]
Here, $I_i$ is the current of the $i$-th afferent synapse, and each synapse has its own time constant $\tau_{s,i}$; $J$ is the neuron-level reset current, whose decay is controlled by $\tau_J$. The states that evolve over time are
\[
h(t)=\left[\begin{array}{c}
I_1(t) \\
\vdots \\
I_M(t) \\
J(t)
\end{array}\right] .
\]
Let the spike train of each input synapse be
\[
x_i(t)=\sum_j\delta(t-t_{ji}),
\]
and the output spike train of the neuron be
\[
z(t)=\sum_j \delta\left(t-t_{j s}\right).
\]
Then the input term is written as
\[
x(t)=\left[\begin{array}{c}
x_1(t) \\
\vdots \\
x_M(t) \\
z(t)
\end{array}\right] .
\]
Thus, HetSyn can be directly written in the following form:
\[
\begin{aligned}
\underbrace{
\begin{bmatrix}
\dot{I}_1 \\
\vdots \\
\dot{I}_M \\
\dot{J}
\end{bmatrix}
}_{\dot{h}}
&=
\underbrace{
\begin{bmatrix}
-\dfrac{1}{\tau_{s,1}} & 0 & \cdots & 0 \\
0 & \ddots & \ddots & \vdots \\
\vdots & \ddots & -\dfrac{1}{\tau_{s,M}} & 0 \\
0 & \cdots & 0 & -\dfrac{1}{\tau_J}
\end{bmatrix}
}_{A}
\underbrace{
\begin{bmatrix}
I_1 \\
\vdots \\
I_M \\
J
\end{bmatrix}
}_{h}
+
\underbrace{
\begin{bmatrix}
w_1 & 0 & \cdots & 0 \\
0 & \ddots & \ddots & \vdots \\
\vdots & \ddots & w_M & 0 \\
0 & \cdots & 0 & \vartheta
\end{bmatrix}
}_{B}
\underbrace{
\begin{bmatrix}
x_1 \\
\vdots \\
x_M \\
z
\end{bmatrix}
}_{x},
\\[10pt]
u(t)
&=
\underbrace{
\begin{bmatrix}
1 \\
\vdots \\
1 \\
-1
\end{bmatrix}^{\mathsf T}
}_{C}
\underbrace{
\begin{bmatrix}
I_1 \\
\vdots \\
I_M \\
J
\end{bmatrix}
}_{h},
\\[10pt]
s(t) &=H\bigl(u(t)-\vartheta\bigr),\qquad h(t^+) =R\bigl(h(t^-),s(t)\bigr)=h(t^-).
\end{aligned}
\]
In HetSyn, the synapse-specific decay coefficients and synaptic weights are learnable, so the decay modes in $A$ and the corresponding input-integration weights in $B$ are adjusted during training. The readout $C=[1,\ldots,1,-1]$ remains fixed, summing the synaptic currents and subtracting the reset current to obtain the membrane potential. Since the spike-triggered update of $J$ is already driven by $z(t)$ through $B$, no additional post-spike update is required and $R$ is the identity in this representation.

From a state-space perspective, HetSyn introduces an independent real-valued decay state for each afferent synapse. Each synaptic time constant $\tau_{s,i}$ determines the decay rate of its corresponding state, allowing different inputs to be retained over different time scales before their contributions are read out as the membrane potential. Its richer temporal dynamics therefore arise from a large set of heterogeneous synaptic decay modes rather than from coupling between states. Despite this fine-grained temporal heterogeneity, once the parameters are learned, the coefficients governing state evolution, input integration, and state readout remain unchanged during temporal processing.
\subsubsection*{7. D-RF \citep{zhang2025drf}}
D-RF starts from the complex oscillatory dynamics of a single RF neuron \citep{izhikevich2001resonate} and extends them to multiple dendritic branches. The $i$-th dendritic branch is represented by a complex state:
\[
z_i(t)=u_i(t)+jv_i(t)\in\mathbb C.
\]
Here, $u_i(t)$ and $v_i(t)$ are the two real-valued components of the resonant state. Its continuous dynamics are
\[
\frac{\mathrm{d} z_i(t)}{\mathrm{d} t}=\left(-\frac{1}{\tau_i}+j \omega_i\right) z_i(t)+\gamma_i I(t),
\]
where $\tau_i$ controls the decay scale of the branch and $\omega_i$ controls its intrinsic oscillation frequency. Meanwhile, D-RF does not use a conventional soft or hard reset, but instead adjusts the threshold according to historical firing activity. The multiple dendritic states form its internal state:
\[
h(t)=\left[\begin{array}{c}
z_1(t) \\
z_2(t) \\
\vdots \\
z_n(t)
\end{array}\right].
\]
Therefore, D-RF can be written in the following form:
\[
\begin{aligned}
\underbrace{
\begin{bmatrix}
\dot{z}_1(t) \\
\dot{z}_2(t) \\
\vdots \\
\dot{z}_n(t)
\end{bmatrix}
}_{\dot{h}(t)}
&=
\underbrace{
\begin{bmatrix}
-\dfrac{1}{\tau_1}+j\omega_1 & 0 & \cdots & 0 \\
0 & -\dfrac{1}{\tau_2}+j\omega_2 & \cdots & 0 \\
\vdots & \vdots & \ddots & \vdots \\
0 & 0 & \cdots & -\dfrac{1}{\tau_n}+j\omega_n
\end{bmatrix}
}_{A}
\underbrace{
\begin{bmatrix}
z_1(t) \\
z_2(t) \\
\vdots \\
z_n(t)
\end{bmatrix}
}_{h(t)}
+
\underbrace{
\begin{bmatrix}
\gamma_1 \\
\gamma_2 \\
\vdots \\
\gamma_n
\end{bmatrix}
}_{B}
x(t),
\\[10pt]
u(t)
&=
\underbrace{
\begin{bmatrix}
c_1 \\
c_2 \\
\vdots \\
c_n
\end{bmatrix}^{\mathsf T}
}_{C}
\underbrace{
\begin{bmatrix}
z_1(t) \\
z_2(t) \\
\vdots \\
z_n(t)
\end{bmatrix}
}_{h(t)},
\\[10pt]
s(t) &=H\bigl(u(t)-\vartheta(t)\bigr),\qquad h(t^+) = h(t^-).
\end{aligned}
\]
In D-RF, the decay rates and oscillation frequencies in $A$, together with the somatic readout weights in $C$, are learnable, whereas the input-integration coefficients in $B$ are not learned. Instead of resetting the dendritic states, D-RF uses a history-dependent adaptive threshold for spike generation, so $R$ is the identity.

From a state-space perspective, D-RF extends the single resonant state of an RF neuron to multiple independently evolving dendritic states with complex eigenvalues $\lambda_i=-1/\tau_i+j\omega_i$. The real part $-1/\tau_i$ determines the decay rate, while the imaginary part $\omega_i$ determines the oscillation frequency. Each branch therefore forms a damped oscillatory mode with a specific time scale and frequency preference. A single RF neuron has a relatively narrow resonance band. D-RF uses branches with different $(\tau_i,\omega_i)$ to establish multiple frequency-selective modes in parallel, then reads out their contributions through $C$ to cover a broader spectrum and multiple time scales. Compared with a temporal basis consisting only of real decay modes, these complex-valued modes represent both decay and oscillation. Despite this richer temporal basis, the coefficients in $A$, $B$, and $C$ remain unchanged during temporal processing.
\subsubsection*{8. PMSN \citep{chen2026pmsn}}
PMSN starts from multi-compartment interaction and divides a neuron into $n$ adjacently coupled compartments. Each compartment has its own membrane leakage and exchanges state information with neighboring compartments. The generalized multi-compartment dynamics first given in the paper are
\[
\begin{aligned}
\dot{v}^{(1)}(t) & =-\frac{1}{\tau_1} v^{(1)}(t)+\beta_{2,1} v^{(2)}(t)+\gamma_1 I(t), \\
\dot{v}^{(2)}(t) & =\beta_{1,2} v^{(1)}(t)-\frac{1}{\tau_2} v^{(2)}(t)+\beta_{3,2} v^{(3)}(t)+\gamma_2 I(t), \\
& \quad \vdots \\
\dot{v}^{(n)}(t) & =\beta_{n-1, n} v^{(n-1)}(t)-\frac{1}{\tau_n} v^{(n)}(t)+\gamma_n I(t), \\
s(t) & =H\left(v^{(n)}(t)-\vartheta\right) .
\end{aligned}
\]
Here, $\tau_i$ controls the leakage time scale of the $i$-th compartment, and $\beta_{i,j}$ controls the coupling between adjacent compartments. The final compartment $v^{(n)}$ serves as the spike-generating output compartment. To enable parallel computation, the feedback from the output compartment to the last hidden compartment is removed, i.e., $\beta_{n,n-1}=0$. Let
\[
h(t)=\left[\begin{array}{c}
v^{(1)}(t) \\
v^{(2)}(t) \\
\vdots \\
v^{(n)}(t)
\end{array}\right].
\]
With $x(t)=I(t)$, PMSN can be written as
\begin{align*}
\underbrace{
\begin{bmatrix}
\dot{v}_1(t) \\
\dot{v}_2(t) \\
\vdots \\
\dot{v}_{n-1}(t) \\
\dot{v}_n(t)
\end{bmatrix}
}_{\dot{h}(t)}
&=
\underbrace{
\begin{bmatrix}
-\dfrac{1}{\tau_1} & \beta_{2,1} & 0 & \cdots & 0 \\
\beta_{1,2} & -\dfrac{1}{\tau_2} & \beta_{3,2} & \cdots & 0 \\
0 & \beta_{2,3} & -\dfrac{1}{\tau_3} & \ddots & \vdots \\
\vdots & \vdots & \ddots & -\dfrac{1}{\tau_{n-1}} & 0 \\
0 & 0 & \cdots & \beta_{n-1,n} & -\dfrac{1}{\tau_n}
\end{bmatrix}
}_{A}
\underbrace{
\begin{bmatrix}
v_1(t) \\
v_2(t) \\
\vdots \\
v_{n-1}(t) \\
v_n(t)
\end{bmatrix}
}_{h(t)}
+
\underbrace{
\begin{bmatrix}
\gamma_1 \\
\gamma_2 \\
\vdots \\
\gamma_{n-1} \\
\gamma_n
\end{bmatrix}
}_{B}
x(t),
\displaybreak[1]\\[10pt]
u(t)
&=
\underbrace{
\begin{bmatrix}
0 \\
0 \\
\vdots \\
0 \\
1
\end{bmatrix}^{\mathsf T}
}_{C}
\underbrace{
\begin{bmatrix}
v_1(t) \\
v_2(t) \\
\vdots \\
v_{n-1}(t) \\
v_n(t)
\end{bmatrix}
}_{h(t)},
\\[10pt]
s(t) &=H\bigl(u(t)-\vartheta\bigr),\qquad h(t^+) =R\bigl(h(t^-),s(t)\bigr).
\end{align*}
After a linear change of basis, PMSN directly parameterizes and learns $A$, $B$, and $C$ in the transformed state space, achieving a fully learnable state-space parameterization. The reset acts only on the final output compartment. Although $A$, $B$, and $C$ are all learnable, once learned, they remain unchanged during temporal processing.

From a state-space perspective, PMSN extends the single state of LIF to an $n$-dimensional coupled state system. The diagonal terms $-1/\tau_i$ describe the intrinsic leakage of different compartments, while the off-diagonal terms $\beta_{i,j}$ couple neighboring compartments and enable continuous information exchange. Although the compartment states and coupling coefficients are real-valued, these interactions can produce complex-conjugate eigenvalues and hence oscillatory modes. After eigendecomposition, the coupled dynamics can be represented by independent modes with different decay rates and oscillation frequencies, forming a multiscale temporal basis.
\subsubsection*{9. TV-MCN}
\begingroup
\setlength{\arraycolsep}{2.5pt}
\allowdisplaybreaks[1]
\begin{align*}
\underbrace{
\begin{bmatrix}
\dot{h}_1(t) \\
\dot{h}_2(t) \\
\vdots \\
\dot{h}_n(t)
\end{bmatrix}
}_{\dot{h}(t)}
&=
\Delta_t
\underbrace{
\begin{bmatrix}
-e^{a_1}+j\omega_1 & 0 & \cdots & 0 \\
0 & -e^{a_2}+j\omega_2 & \cdots & 0 \\
\vdots & \vdots & \ddots & \vdots \\
0 & 0 & \cdots & -e^{a_n}+j\omega_n
\end{bmatrix}
}_{A}
\underbrace{
\begin{bmatrix}
h_1(t) \\
h_2(t) \\
\vdots \\
h_n(t)
\end{bmatrix}
}_{h(t)}+
\Delta_t
\underbrace{
\begin{bmatrix}
b_{1,t}^{\mathrm{R}}+j b_{1,t}^{\mathrm{I}} \\
b_{2,t}^{\mathrm{R}}+j b_{2,t}^{\mathrm{I}} \\
\vdots \\
b_{n,t}^{\mathrm{R}}+j b_{n,t}^{\mathrm{I}}
\end{bmatrix}
}_{B_t}
x(t),
\\[10pt]
u(t)
&=
\underbrace{
\begin{bmatrix}
c_{1,t}^{\mathrm{R}}+j c_{1,t}^{\mathrm{I}} \\
c_{2,t}^{\mathrm{R}}+j c_{2,t}^{\mathrm{I}} \\
\vdots \\
c_{n,t}^{\mathrm{R}}+j c_{n,t}^{\mathrm{I}}
\end{bmatrix}^{\mathsf T}
}_{C_t}
\underbrace{
\begin{bmatrix}
h_1(t) \\
h_2(t) \\
\vdots \\
h_n(t)
\end{bmatrix}
}_{h(t)},
\\[10pt]
s(t) &= H\bigl(u(t)-\vartheta\bigr),\qquad h\left(t^{+}\right) = R\left(h\left(t^{-}\right),s(t)\right).
\end{align*}
\endgroup
where
\[
\begin{aligned}
\Delta_t &= \operatorname{softplus} \left( b_\Delta+W_\Delta q_t \right),\\
q_t&=\operatorname{SiLU}(W_q \mathbf{x}_t),\\
B_t&=W_B \mathbf{x}_t,\\
C_t^{\mathsf T}&=W_C \mathbf{x}_t..
\end{aligned}
\]
The superscripts $\mathrm{R}$ and $\mathrm{I}$ denote real and imaginary parts, respectively.

\subsection{Neuron Models with Nonlinear State Evolution}

\subsubsection*{1. Hodgkin--Huxley (HH) \citep{hodgkin1952quantitative}}
Starting from membrane capacitance and sodium, potassium, and leak currents, HH describes action-potential generation and recovery through changes in ionic conductances with membrane potential. In addition to the membrane potential $V$, the model contains a sodium-channel activation variable $m$, a sodium-channel inactivation variable $q$, and a potassium-channel activation variable $n$. Here, the original inactivation variable $h$ is denoted by $q$ to distinguish it from the state vector in the framework. The continuous dynamics are
\[
\begin{aligned}
C_m\dot V
&=I-\bar g_{\mathrm{Na}}m^3q(V-E_{\mathrm{Na}})
-\bar g_{\mathrm K}n^4(V-E_{\mathrm K})-g_{\mathrm L}(V-E_{\mathrm L}),\\
\dot m&=\alpha_m(V)(1-m)-\beta_m(V)m,\\
\dot q&=\alpha_q(V)(1-q)-\beta_q(V)q,\\
\dot n&=\alpha_n(V)(1-n)-\beta_n(V)n.
\end{aligned}
\]
Here, $C_m$ is the membrane capacitance; $\bar g_{\mathrm{Na}},\bar g_{\mathrm K},g_{\mathrm L}$ are conductance parameters; and $E_{\mathrm{Na}},E_{\mathrm K},E_{\mathrm L}$ are the corresponding reversal potentials. The gating transition rates $\alpha$ and $\beta$ depend on the current membrane potential.

The membrane potential and the three gating variables all evolve over time. We define the state and input as
\[
h(t)=\begin{bmatrix}V(t)\\m(t)\\q(t)\\n(t)\\1\end{bmatrix},
\qquad x(t)=I(t).
\]
The final coordinate is fixed at $1$. The membrane-potential equation can be written as
\[
\dot V=-a_VV+b_V+\frac{I}{C_m},
\]
where
\[
\begin{aligned}
a_V&=\frac{\bar g_{\mathrm{Na}}m^3q+\bar g_{\mathrm K}n^4+g_{\mathrm L}}{C_m},\\
b_V&=\frac{\bar g_{\mathrm{Na}}m^3qE_{\mathrm{Na}}+\bar g_{\mathrm K}n^4E_{\mathrm K}+g_{\mathrm L}E_{\mathrm L}}{C_m}.
\end{aligned}
\]
Each gating equation can likewise be written as $\dot z=-(\alpha_z+\beta_z)z+\alpha_z$, where $z\in\{m,q,n\}$. HH can therefore be written in the following framework form:
\[
\begin{aligned}
\underbrace{\begin{bmatrix}\dot V\\\dot m\\\dot q\\\dot n\\0\end{bmatrix}}_{\dot h(t)}
&=
\underbrace{
\begin{bmatrix}
-a_V&0&0&0&b_V\\
0&-(\alpha_m+\beta_m)&0&0&\alpha_m\\
0&0&-(\alpha_q+\beta_q)&0&\alpha_q\\
0&0&0&-(\alpha_n+\beta_n)&\alpha_n\\
0&0&0&0&0
\end{bmatrix}}_{A(h)}
\underbrace{\begin{bmatrix}V\\m\\q\\n\\1\end{bmatrix}}_{h(t)}
+
\underbrace{\begin{bmatrix}C_m^{-1}\\0\\0\\0\\0\end{bmatrix}}_B x(t),\\[10pt]
u(t)&=
\underbrace{\begin{bmatrix}1&0&0&0&0\end{bmatrix}}_C
\underbrace{\begin{bmatrix}V\\m\\q\\n\\1\end{bmatrix}}_{h(t)}=V(t),\\[10pt]
s(t)&=H_{\uparrow}\!\left(u(t)-\vartheta\right),\qquad
h(t^+)=R\!\left(h(t^-),s(t)\right)=h(t^-).
\end{aligned}
\]
Here, the gating rates $\alpha$ and $\beta$ are evaluated at the current membrane potential. $H_{\uparrow}$ records a spike when the membrane potential crosses the detection threshold from below. Action potentials and repolarization in HH arise naturally from the continuous dynamics, so $R$ is the identity mapping here.

From a state-space perspective, HH consists of the membrane potential and three gating states. The gating states determine the coefficients governing membrane-potential evolution, while the membrane potential in turn affects the evolution of the gating states. Thus, the state-evolution matrix $A$ is a function of the state $h$. Unlike the constant-coefficient decay in LIF, both the effective evolution rates and the values toward which individual states evolve depend on other state variables. By adding states that regulate one another, HH allows the neuron's time scales and state evolution to change with its own state. The input integration $B$ and state readout $C$ remain unchanged, respectively injecting current into the membrane potential and reading it out.

\subsubsection*{2. Izhikevich Neuron \citep{izhikevich2003simple}}
The Izhikevich model describes a neuron using the membrane potential and a recovery variable. A quadratic voltage term produces rapid depolarization, while the recovery variable provides negative feedback. Here, the original recovery variable $u$ is denoted by $w$ to distinguish it from the membrane-potential readout in the framework. The dynamics are
\[
\begin{aligned}
\dot v&=0.04v^2+5v+140-w+I,\\
\dot w&=a(bv-w).
\end{aligned}
\]
When $v\geq30\,\mathrm{mV}$, a spike is recorded and the following updates are applied:
\[
v\leftarrow c,\qquad w\leftarrow w+d.
\]
Here, $a$ controls the rate of change of the recovery variable, $b$ controls its response to the membrane potential, $c$ is the post-spike membrane potential, and $d$ is the increment in the recovery variable caused by each spike.

The membrane potential $v$ and recovery variable $w$ jointly form the internal state. We define the state and input as
\[
h(t)=\begin{bmatrix}v(t)\\w(t)\\1\end{bmatrix},\qquad x(t)=I(t).
\]
Writing the voltage term as $0.04v^2+5v=(0.04v+5)v$ gives the following framework form:
\[
\begin{aligned}
\underbrace{\begin{bmatrix}\dot v\\\dot w\\0\end{bmatrix}}_{\dot h(t)}
&=
\underbrace{
\begin{bmatrix}
0.04v+5&-1&140\\
ab&-a&0\\
0&0&0
\end{bmatrix}}_{A(h)}
\underbrace{\begin{bmatrix}v\\w\\1\end{bmatrix}}_{h(t)}
+\underbrace{\begin{bmatrix}1\\0\\0\end{bmatrix}}_B x(t),\\[10pt]
u(t)&=
\underbrace{\begin{bmatrix}1&0&0\end{bmatrix}}_C
\underbrace{\begin{bmatrix}v\\w\\1\end{bmatrix}}_{h(t)}=v(t),\\[10pt]
s(t)&=H\!\left(u(t)-30\right),\qquad
h(t^+)=R\!\left(h(t^-),s(t)\right)
=\begin{bmatrix}(1-s)v^-+sc\\w^-+sd\\1\end{bmatrix}.
\end{aligned}
\]
Here, $B$ integrates the input current into the membrane-potential state, and $C$ selects this state as the membrane-potential readout. After a spike, $R$ resets $v$ to $c$ and increases $w$ by $d$.

From a state-space perspective, the Izhikevich model is a two-dimensional coupled system. The diagonal term \(0.04v+5\) in \(A(h)\), corresponding to the membrane potential, varies with the state. The membrane potential can therefore alter the rule governing its own evolution, making the model state-dependent and nonlinear. However, the decay coefficient of the recovery state and the coupling coefficients between the two states remain fixed and do not depend on the states themselves. The input integration matrix \(B\) and the readout matrix \(C\) also remain unchanged.

\subsubsection*{3. DGN \citep{bai2025gating}}
DGN adds a set of synaptic current states alongside the membrane potential. These states both drive the membrane potential and regulate its decay rate through dynamic conductance. Its continuous dynamics are
\[
\begin{gathered}
\tau_s \frac{d D_i(t)}{d t}=-D_i(t)+z_i(t) \\
\frac{d V(t)}{d t}=-\left(g_l+\sum_{i=1}^N C_i D_i(t)\right) V(t)+\sum_{i=1}^N W_i D_i(t).
\end{gathered}
\]
Here, $D_i$ is the time-decaying synaptic current state of the $i$-th synapse, $C_i$ is the conductance weight that aggregates these states, and $W_i$ is the input-current weight. The internal state should therefore include all synaptic current states and the membrane potential:
\[
h(t)=\left[\begin{array}{c}
D_1(t) \\
D_2(t) \\
\vdots \\
D_N(t) \\
V(t)
\end{array}\right],
\]
and the input consists of all presynaptic spikes:
\[
x(t)=\left[\begin{array}{c}
z_1(t) \\
z_2(t) \\
\vdots \\
z_N(t)
\end{array}\right] .
\]
DGN can then be written as
\begingroup
\setlength{\arraycolsep}{2pt}
\[
\resizebox{\linewidth}{!}{$\displaystyle
\begin{aligned}
\underbrace{
\begin{bmatrix}
\dot{D}_1(t) \\
\dot{D}_2(t) \\
\vdots \\
\dot{D}_N(t) \\
\dot{V}(t)
\end{bmatrix}
}_{\dot{h}(t)}
&=
\underbrace{
\begin{bmatrix}
-\dfrac{1}{\tau_s} & 0 & \cdots & 0 & 0 \\
0 & -\dfrac{1}{\tau_s} & \cdots & 0 & 0 \\
\vdots & \vdots & \ddots & \vdots & \vdots \\
0 & 0 & \cdots & -\dfrac{1}{\tau_s} & 0 \\
w_1 & w_2 & \cdots & w_N &
-\left(g_l+\displaystyle\sum_{i=1}^{N}c_iD_i(t)\right)
\end{bmatrix}
}_{A(h)}
\underbrace{
\begin{bmatrix}
D_1(t) \\
D_2(t) \\
\vdots \\
D_N(t) \\
V(t)
\end{bmatrix}
}_{h(t)}
+
\underbrace{
\begin{bmatrix}
\dfrac{1}{\tau_s} & 0 & \cdots & 0 \\
0 & \dfrac{1}{\tau_s} & \cdots & 0 \\
\vdots & \vdots & \ddots & \vdots \\
0 & 0 & \cdots & \dfrac{1}{\tau_s} \\
0 & 0 & \cdots & 0
\end{bmatrix}
}_{B}
\underbrace{
\begin{bmatrix}
z_1(t) \\
z_2(t) \\
\vdots \\
z_N(t)
\end{bmatrix}
}_{x(t)},
\end{aligned}
$}
\]
\endgroup

\[
\begin{aligned}
u(t)
&=
\underbrace{
\begin{bmatrix}
0 \\
\vdots \\
0 \\
1
\end{bmatrix}^{\mathsf T}
}_{C}
\underbrace{
\begin{bmatrix}
D_1(t) \\
\vdots \\
D_N(t) \\
V(t)
\end{bmatrix}
}_{h(t)},
\\[10pt]
s(t) &=H\bigl(u(t)-\vartheta\bigr),\qquad h(t^+) =R\bigl(h(t^-),s(t)\bigr).
\end{aligned}
\]
From a state-space perspective, DGN introduces state-dependent evolution in the membrane-potential dimension. Rather than being determined directly by the current input, its state evolution depends on input history stored in the synaptic states $D_i(t)$, whose weighted combination changes the decay coefficient of the membrane potential $V$. The state matrix should therefore be viewed as $A(h)$. DGN thus forms a nonlinear state system, and its state updates over the full sequence cannot be computed directly in parallel using a fixed convolution or an affine scan. Besides, this dynamic modulation is confined to membrane-potential decay, while the remaining synaptic states still evolve with fixed time constants. Meanwhile, $B$ writes inputs into the synaptic states with the fixed coefficient $\tau_s^{-1}$, and $C$ reads out the membrane potential with a fixed mapping.

\section{Experimental Settings}
\label{app:experimental_settings}

\subsection{Comparison Settings and Task-Level Training Configurations}
\label{app:comparison_training_settings}

The experiments include shared-network neuron comparisons and complete-network comparisons. The shared-network comparisons in Tables~\ref{tab:neuron_comparison} and~\ref{tab:context_neuron_comparison} use a fully connected network with two layers of 128 neurons for all nine neuron models. The network does not use cross-layer accumulated spike residuals, and the task head reads the activity of the final layer. Input encoders and task heads are configured for each dataset and kept identical across neuron models within the same task. Each method retains its own state variables, synaptic integration, and spike-generation mechanism; DH-LIF, HetSyn, and DGN retain their internal synaptic projections and states.

In the complete-network comparison in Table~\ref{tab:sota_comparison}, each method uses its representative architecture. Each method retains the same network across the six synthetic sequence tasks, with only the input and output layers adjusted. All methods use the same data splits and test protocols.

Table~\ref{tab:shared_training_configurations} lists the task-level training configurations for the shared-network comparisons. Learning rates refer to global base learning rates; method-specific settings are listed separately below. The complete-network comparison uses the same preset training budgets and batch sizes for the six synthetic sequence tasks, with learning rates, optimizers, and architectures specified in Section~\ref{app:full_network_configurations}.

\begin{table}[!htbp]
\centering
\caption{Task-level training configurations for the shared-network comparisons.}
\label{tab:shared_training_configurations}
\footnotesize
\setlength{\tabcolsep}{3pt}
\renewcommand{\arraystretch}{1.15}
\renewcommand{\tabularxcolumn}[1]{m{#1}}
\begin{tabularx}{\linewidth}{@{}lcccc>{\centering\arraybackslash}X@{}}
\toprule
\makecell[l]{Task} & \makecell[c]{Preset training\\budget} & \makecell[c]{Batch\\size} & \makecell[c]{Base learning\\rate} & Optimizer & Normalization \\
\midrule
GSC & 200 epochs & 128 & 0.001 & AdamW & LayerNorm \\
SSC & 128 epochs & 128 & 0.001 & Adam & LayerNorm \\
SHD & 100 epochs & 40 & 0.01 & AdamW & LayerNorm \\
DVS128 Gesture & 200 epochs & 32 & 0.005 & AdamW & Masked BatchNorm \\
MAD ICR & 200 epochs & 128 & 0.0005 & AdamW & LayerNorm \\
MAD Noisy ICR & 200 epochs & 128 & 0.0005 & AdamW & LayerNorm \\
Selective Copying & 40,000 steps & 64 & 0.0001 & Adam & LayerNorm \\
Adding & 100,000 steps & 128 & 0.001 & AdamW & None \\
STORE--RECALL & 4,000 steps & 32 & 0.001 & Adam & LayerNorm \\
Induction & 204,800 steps & 8 & 0.001 & Adam & BatchNorm \\
\bottomrule
\end{tabularx}
\end{table}

GSC uses a step learning-rate schedule, multiplying the learning rate by 0.1 every 30 epochs. SSC, Selective Copying, and Induction use constant learning rates.

Model selection uses the checkpoint with the best validation performance, measured by the lowest MSE for Adding and the highest task-specific accuracy for the other tasks. Learning rates are adjusted for individual tasks where needed, with the adopted values reported in the configuration tables.

The common linear, embedding, and normalization modules are initialized through their respective \texttt{reset\_parameters()} methods using a separate shared random stream. Internal neuron parameters follow the initialization scheme of each method.

\subsection{Internal State Sizes, Initialization, and Task-Specific Settings}
\label{app:neuron_initialization_settings}

Table~\ref{tab:neuron_initialization} summarizes the internal state sizes and baseline initialization schemes for the shared-network comparisons. Table~\ref{tab:neuron_task_specific_settings} lists the adopted task-specific adjustments; parameters not listed there retain their baseline settings. Here, $\mathcal{N}(\mu,\sigma^2)$ denotes a normal distribution with mean $\mu$ and standard deviation $\sigma$, and $\mathcal{U}(a,b)$ denotes a uniform distribution over $[a,b]$.

\begin{table}[!htbp]
\centering
\caption{Internal state sizes and baseline initialization schemes of the neuron models.}
\label{tab:neuron_initialization}
\footnotesize
\setlength{\tabcolsep}{5pt}
\renewcommand{\arraystretch}{1.15}
\renewcommand{\tabularxcolumn}[1]{m{#1}}
\begin{tabularx}{\linewidth}{@{}>{\centering\arraybackslash}m{0.16\linewidth}>{\raggedright\arraybackslash}X@{}}
\toprule
Neuron & Internal state size and initialization \\
\midrule
LIF & The decay coefficient is 0.9, and the firing threshold is 1. \\
\addlinespace[3pt]
ad-LIF & Membrane time constants are initialized from $\mathcal{N}(20,5^2)$, and adaptation time constants from $\mathcal{N}(200,5^2)$. The base threshold is 1.6. \\
\addlinespace[3pt]
TC-LIF & The two raw coupling parameters are initialized to 0, corresponding to an effective value of 0.5 after the sigmoid transformation. The firing threshold is 1, and the dendritic reset coefficient is 0.5. \\
\addlinespace[3pt]
DH-LIF & Four dendritic branches are used. Membrane parameters are initialized from $\mathcal{U}(0,4)$ and branch parameters from $\mathcal{U}(2,6)$. The firing threshold is 1. \\
\addlinespace[3pt]
HetSyn & Synaptic weights are initialized from $\mathcal{N}(0,0.01^2)$. Synaptic time constants are initialized to 20 ms, the time step is 1 ms, and the firing threshold is 1. \\
\addlinespace[3pt]
DGN & The time step is 1, and the synaptic time constant is 20. Conductance and current weights are both initialized from $\mathcal{N}(0.01,0.005^2)$. \\
\addlinespace[3pt]
D-RF & A four-dimensional complex state is used. The initial decay rate is 0.5, and frequencies are set to $n\pi$. Time steps are initialized log-uniformly over $[0.001,0.1]$, and complex-valued readout weights are randomly initialized. \\
\addlinespace[3pt]
PMSN & Four compartments are used with Coupled-Spectrum initialization. Input integration and state readout use Structured-Physical $B$ and Structured-Modal $C$ initialization, respectively. \\
\addlinespace[3pt]
TV-MCN & Four compartments are used with a conditioning rank of 32. Initial decay rates are $1,2,3,4$, and frequencies are $0,\pi,2\pi,3\pi$. Internal time scales are initialized log-uniformly over $[0.001,0.1]$. Controller modules use their respective Xavier or uniform initialization schemes. \\
\bottomrule
\end{tabularx}
\end{table}

For the baseline neurons, we take the initialization schemes of the respective original methods as starting points; the baseline settings used here are summarized in Table~\ref{tab:neuron_initialization}. We then adjust selected initialization settings and neuron parameters for individual tasks, while keeping the depth, width, and connectivity of the shared network unchanged. Table~\ref{tab:neuron_task_specific_settings} lists the adopted adjustments, including task-specific learning rates.

\begin{table}[!htbp]
\centering
\caption{Method- and task-specific configurations for the shared-network comparisons.}
\label{tab:neuron_task_specific_settings}
\footnotesize
\setlength{\tabcolsep}{5pt}
\renewcommand{\arraystretch}{1.15}
\renewcommand{\tabularxcolumn}[1]{m{#1}}
\begin{tabularx}{\linewidth}{@{}>{\centering\arraybackslash}m{0.28\linewidth}>{\raggedright\arraybackslash}X@{}}
\toprule
Neuron and task & Adopted configuration \\
\midrule
LIF -- STORE--RECALL & The decay coefficient is changed from 0.9 to 0.99. \\
\addlinespace[3pt]
ad-LIF -- SSC & Membrane time constants are initialized from $\mathcal{N}(5,1.25^2)$. \\
\addlinespace[3pt]
ad-LIF -- Adding & Membrane time constants are initialized from $\mathcal{N}(5,1^2)$, and adaptation time constants from $\mathcal{N}(200,50^2)$. The base threshold is 0.4. \\
\addlinespace[3pt]
ad-LIF -- Induction & Membrane time constants are initialized from $\mathcal{N}(5,1.25^2)$, and adaptation time constants from $\mathcal{N}(200,50^2)$. The base threshold is 0.4. \\
\addlinespace[3pt]
ad-LIF -- STORE--RECALL & Membrane time constants are initialized from $\mathcal{N}(5,1.25^2)$, and adaptation time constants from $\mathcal{N}(200,50^2)$. The base threshold remains 1.6. \\
\addlinespace[3pt]
TC-LIF -- STORE--RECALL & The initial effective coupling value is changed from 0.5 to 0.125. \\
\addlinespace[3pt]
DGN -- STORE--RECALL & Conductance weights are initialized from $\mathcal{N}(0.001,0.0005^2)$; current weights retain the initialization $\mathcal{N}(0.01,0.005^2)$. \\
\addlinespace[3pt]
PMSN -- SHD & The global base learning rate is 0.001. \\
\addlinespace[3pt]
TV-MCN -- SHD & The global base learning rate is 0.002. \\
\addlinespace[3pt]
TV-MCN -- MAD Noisy ICR & The global base learning rate is 0.001. \\
\bottomrule
\end{tabularx}
\end{table}

\subsection{Architectures and Training Configurations for Complete-Network Comparisons}
\label{app:full_network_configurations}

The complete-network comparison includes PMSN, D-RF, SpikingSSMs, S5-RF, Spiking LMUFormer, and TV-MCN. Their architectures are listed in Table~\ref{tab:full_network_architectures}. Each network retains its own normalization, residual connections, and internal memory structure. The normalization settings in Table~\ref{tab:shared_training_configurations} apply only to the shared-network comparisons.

\begin{table}[!htbp]
\centering
\caption{Architectures used in the complete-network comparison.}
\label{tab:full_network_architectures}
\footnotesize
\setlength{\tabcolsep}{5pt}
\renewcommand{\arraystretch}{1.15}
\renewcommand{\tabularxcolumn}[1]{m{#1}}
\begin{tabularx}{\linewidth}{@{}>{\centering\arraybackslash}m{0.22\linewidth}>{\raggedright\arraybackslash}X@{}}
\toprule
Method & Architecture \\
\midrule
PMSN & Two stages, each containing three spiking layers, with widths of 128 and 256, respectively. Each neuron has four compartments. Average pooling is applied along the feature dimension between stages and at the network output. \\
\addlinespace[3pt]
D-RF & Six independent residual blocks of width 512, with a complex-state dimension of 8 and a causal convolution at the input. Following the adopted forward implementation, the first block is executed twice, giving six independent sets of block parameters and seven block calls. \\
\addlinespace[3pt]
SpikingSSMs & Sixteen layers of width 1024 and state dimension 64, with a pre-LayerNorm residual structure. The authors' pretrained SDN modules are loaded and frozen. \\
\addlinespace[3pt]
S5-RF & Four layers of width 512, with HiPPO spectral initialization divided into 16 blocks. The RF core and LI readout are retained. \\
\addlinespace[3pt]
Spiking LMUFormer & One native block with hidden width 256, memory size 512, a feedforward expansion factor of 4, and memory time-scale parameter $\theta=128$. \\
\addlinespace[3pt]
TV-MCN & Two layers of 128 neurons, with four compartments per neuron. Each layer includes a depthwise causal convolution with kernel size 4. The convolutions are initialized as identity mappings and add 1,280 parameters in total. \\
\bottomrule
\end{tabularx}
\end{table}

PMSN retains its coupled-spectrum and structured $B/C$ initialization. Ordinary dropout is set to 0 on Selective Copying and STORE--RECALL; the associated spike dropout is also set to 0 on STORE--RECALL. D-RF retains its complex-pole, random complex-readout, and log-uniform time-scale initialization. S5-RF initializes its internal step sizes log-uniformly over $[0.004,0.1]$. Spiking LMUFormer uses a fixed Legendre memory system; its native linear weights use truncated normal initialization with a standard deviation of 0.02, and biases are initialized to 0. TV-MCN uses the same neuron initialization as in the shared-network experiments, with identity initialization for the added convolutions.

For the complete-network comparison, we take the learning rates from each baseline's original configuration as a starting point and make task-specific adjustments where needed. Table~\ref{tab:full_network_learning_rates} lists the adopted global base learning rates for all models. Spiking LMUFormer uses Adam for all tasks. The other five methods use AdamW for MAD ICR, MAD Noisy ICR, and Adding, and Adam for Selective Copying, STORE--RECALL, and Induction. On Induction, PMSN uses the task-uniform Adam configuration with a learning rate of 0.001 and weight decay of 0.

\begin{table}[!htbp]
\centering
\caption{Global base learning rates for the complete-network comparison.}
\label{tab:full_network_learning_rates}
\footnotesize
\setlength{\tabcolsep}{2.5pt}
\renewcommand{\arraystretch}{1.15}
\begin{tabular*}{\linewidth}{@{\extracolsep{\fill}}ccccccc@{}}
\toprule
\makecell[c]{Method} & \makecell[c]{MAD ICR} & \makecell[c]{MAD Noisy\\ICR} & \makecell[c]{Selective\\Copying} & \makecell[c]{STORE--\\RECALL} & \makecell[c]{Induction} & \makecell[c]{Adding} \\
\midrule
PMSN & 0.01 & 0.01 & 0.01 & 0.01 & 0.001 & 0.01 \\
D-RF & 0.005 & 0.005 & 0.005 & 0.005 & 0.005 & 0.005 \\
SpikingSSMs & 0.0005 & 0.0005 & 0.0005 & 0.0005 & 0.0005 & 0.0005 \\
S5-RF & 0.004 & 0.004 & 0.004 & 0.004 & 0.004 & 0.004 \\
Spiking LMUFormer & 0.0025 & 0.0025 & 0.0025 & 0.0025 & 0.0025 & 0.0025 \\
TV-MCN & 0.0005 & 0.001 & 0.0001 & 0.001 & 0.001 & 0.001 \\
\bottomrule
\end{tabular*}
\end{table}

\subsection{Evaluation Protocol}
All methods use the same data splits and test protocols.

Adding reports mean squared error (MSE) at sequence length $L=1000$. Tables~\ref{tab:context_neuron_comparison} and~\ref{tab:sota_comparison} report MSE in units of $10^{-1}$ and $10^{-2}$, respectively, rounded to four decimal places after rescaling. For MAD and Selective Copying, token accuracy measures the fraction of target tokens predicted correctly. For 20D STORE--RECALL, bit accuracy measures the fraction of correctly recalled bits.

For Induction, models are trained at sequence length $L=256$ and evaluated without retraining at $L\in\{64,128,256,512,1024,2048,4096\}$. We report the equally weighted mean accuracy across these seven test lengths and present the accuracy at each length separately.

For the Adding visualization in Figure~\ref{fig:adding_predictions}, all nine neuron models use the shared fully connected network and the checkpoints corresponding to their results in Table~\ref{tab:context_neuron_comparison}. All panels use the same 512 test sequences at $L=1000$ and identical axis limits. The dashed line denotes $y=x$. The MSE values in the table are computed over the full test set.

For MAD Noisy ICR, the main comparison uses sequence length $L=128$ and irrelevant-input proportion $\rho=0.2$. In Figure~\ref{fig:induction_noise} (right), each model is trained at $\rho=0.2$, and the same checkpoint is evaluated at $\rho\in\{0.2,0.4,0.6,0.8\}$ without retraining or checkpoint reselection. Each setting contains 1,280 test sequences, with identical inputs and targets across models. Token accuracy is pooled over supervised recall positions. Data generation follows the MAD protocol at each noise fraction, so the number of content pairs and supervised recall positions also vary with $\rho$.

For the delayed-recall experiment in Figure~\ref{fig:sc_ablation_recall} (right), we evaluate the checkpoints corresponding to Table~\ref{tab:ablation} on the same 128 test sequences of base length $L=4096$. We insert $0,256,512,1024,$ or $2048$ irrelevant symbols immediately before the recall delimiter, preserving the original input prefix and the 16-token target sequence. All checkpoints are evaluated without retraining.

\subsection{Parameter and Computational Cost Estimation}
\label{app:computational_cost}

For Table~\ref{tab:sota_comparison}, we report whole-network parameter counts (M) and estimated FLOPs (G/sequence), including the input encoder, network backbone, and task head. Both trainable and frozen model parameters are counted, shared parameters are counted once, and each complex-valued parameter is counted as two real-valued scalars. Non-parameter buffers are excluded. For Induction, the reported cost is the equally weighted average over the seven evaluation lengths after next-token shifting.

FLOPs are analytical estimates of dense arithmetic for a single-sequence forward pass with batch size 1 in evaluation mode. Spike sparsity does not reduce the count. A real multiplication and addition each count as one operation, so a real multiply--accumulate counts as two operations. A complex multiplication and addition count as six and two real operations, respectively. For FFT-based computation, a real FFT of length $n$ is estimated as $2.5n\log_2 n$ operations and a complex FFT as $5n\log_2 n$ operations per transform. One-time initialization, embedding lookups, indexing, comparisons, data movement, and backward computation are excluded. These FLOPs are therefore analytical arithmetic estimates rather than measurements of hardware latency, energy, or instruction count.

PMSN, D-RF, SpikingSSMs, and S5-RF are counted by tracing the arithmetic operators executed in their forward passes. This includes forward-pass kernel generation and, for SpikingSSMs, the frozen SDN modules. TV-MCN is counted using its reference recurrence, including the causal convolutions, input-dependent generation of $\Delta_t$, $B_t$, and $C_t$, discretization, and state updates. Spiking LMUFormer is estimated from its modules using precomputed kernel spectra; the cost of constructing these spectra is not included, and each LIF element is assigned six arithmetic operations.

The counters differ slightly in the bookkeeping of normalization and FFT scaling. For PMSN, D-RF, SpikingSSMs, and S5-RF, an affine LayerNorm applied to a tensor with $m$ elements arranged into $r$ independently normalized rows is counted as $7m+2r$ operations. An inference-mode BatchNorm with $m$ elements and $c$ channels is counted as $2m+5c$ operations. These counters also include FFT normalization scaling separately. For TV-MCN and Spiking LMUFormer, LayerNorm and BatchNorm are counted as $7m$ and $2m$ operations, respectively, without separate FFT-scaling terms.

\subsection{Parameter Counts of the Shared Network}
\label{app:shared_network_parameters}

Table~\ref{tab:shared_network_parameters} reports the total trainable parameter counts of the two-layer, 128-neuron-per-layer networks used in Tables~\ref{tab:neuron_comparison} and~\ref{tab:context_neuron_comparison}. Counts include the fully connected projections and neuron-specific parameters, but exclude the input encoder, normalization layers, and task head. Each complex-valued parameter counts as two real-valued scalars. 

\begin{table}[!htbp]
\centering
\caption{Trainable parameter counts of the shared two-layer network. K denotes $10^3$ parameters. \textbf{Bold} and \protect\uline{underlining} mark the smallest and second-smallest counts, respectively.}
\label{tab:shared_network_parameters}
\footnotesize
\setlength{\tabcolsep}{16pt}
\renewcommand{\arraystretch}{1.15}
\begin{tabular}{cc}
\toprule
Neuron & Trainable Params (K) \\
\midrule
LIF \citep{burkitt2006review} & \textbf{33.024} \\
ad-LIF \citep{baronig2025advancing} & 33.536 \\
TC-LIF \citep{zhang2024tclif} & \uline{33.028} \\
DH-LIF \citep{zheng2024temporal} & 133.376 \\
HetSyn \citep{deng2025hetsyn} & 65.536 \\
DGN \citep{bai2025gating} & 65.536 \\
D-RF \citep{zhang2025drf} & 37.632 \\
PMSN \citep{chen2026pmsn} & 39.680 \\
\midrule
TV-MCN & 56.064 \\
\bottomrule
\end{tabular}
\end{table}

Relative to time-invariant multi-compartment designs, TV-MCN introduces additional input-dependent projections to generate $\Delta_t$, $B_t$, and $C_t$, increasing its parameter count over D-RF and PMSN. Its count remains below those of DH-LIF, HetSyn, and DGN in this configuration. DH-LIF uses a branch-expanded input projection for its four dendritic branches, together with membrane and branch time-constant parameters; the reported count includes all entries of the dense projection tensor, including masked-out entries. HetSyn assigns a learnable weight and decay coefficient to each input--neuron pair. DGN maintains a trace for each presynaptic input and uses separate conductance and current weight matrices to map these traces to each neuron.

\section{Additional Recall Results}
\label{app:additional_recall_results}

Exact-sequence accuracy measures the fraction of examples in which all target tokens are predicted correctly for Selective Copying, MAD ICR, and MAD Noisy ICR. For 20D STORE--RECALL, word accuracy measures the fraction of recalled words with all 20 bits correct. These results use the same checkpoints and test sets as the corresponding main-text tables.

\subsection{Comparison with Existing Spiking Neurons}

Table~\ref{tab:neuron_complete_recall} reports supplementary exact-sequence and word accuracies for the shared-network comparison in Table~\ref{tab:context_neuron_comparison}. Many baseline neurons obtain zero or near-zero exact-sequence accuracy, indicating that correct predictions at individual positions rarely extend to complete target-sequence recovery. Even high token or bit accuracy can conceal incomplete recall: HetSyn achieves 97.44\% token accuracy on MAD ICR but only 24.38\% exact-sequence accuracy, while DH-LIF achieves 99.27\% bit accuracy on STORE--RECALL but 86.27\% word accuracy. These stricter metrics reveal limitations in recovering complete sequences or words that are less apparent from per-position accuracy. TV-MCN reaches 99.92\% and 99.38\% exact-sequence accuracy on MAD ICR and MAD Noisy ICR, respectively, and 98.40\% word accuracy on STORE--RECALL.

\begin{table}[!htbp]
\centering
\caption{Exact-sequence and word accuracies for the shared-network neuron comparison. All values are in \%. \textbf{Bold} indicates the best results, and \protect\uline{underlining} marks nonzero second-best results.}
\label{tab:neuron_complete_recall}
\footnotesize
\setlength{\tabcolsep}{3.0pt}
\renewcommand{\arraystretch}{1.15}
\begin{tabular*}{\linewidth}{@{\extracolsep{\fill}}ccccc@{}}
\toprule
\makecell[c]{Neuron}
& \makecell[c]{Sel. Copy.\\Exact}
& \makecell[c]{MAD ICR\\Exact}
& \makecell[c]{MAD Noisy\\Exact}
& \makecell[c]{STORE--RECALL\\Word} \\
\midrule
LIF \citep{burkitt2006review} & 0.00 & 0.00 & 0.00 & 0.00 \\
ad-LIF \citep{baronig2025advancing} & 0.00 & 0.00 & 0.00 & 0.00 \\
TC-LIF \citep{zhang2024tclif} & 0.00 & 0.00 & 0.00 & 0.00 \\
DH-LIF \citep{zheng2024temporal} & 0.00 & 0.00 & 0.00 & \uline{86.27} \\
HetSyn \citep{deng2025hetsyn} & 0.00 & \uline{24.38} & \uline{0.31} & 0.00 \\
DGN \citep{bai2025gating} & 0.00 & 0.00 & 0.00 & 0.00 \\
D-RF \citep{zhang2025drf} & 0.00 & 0.00 & 0.00 & 78.79 \\
PMSN \citep{chen2026pmsn} & 0.00 & 0.00 & 0.00 & 26.02 \\
\midrule
TV-MCN & \textbf{37.72} & \textbf{99.92} & \textbf{99.38} & \textbf{98.40} \\
\bottomrule
\end{tabular*}
\end{table}

\subsection{Comparison with State-of-the-Art Methods}

Table~\ref{tab:network_complete_recall} reports supplementary exact-sequence and word accuracies for the complete-network comparison in Table~\ref{tab:sota_comparison}.

\begin{table}[!htbp]
\centering
\caption{Exact-sequence and word accuracies for the complete-network comparison. All values are in \%. \textbf{Bold} indicates the best results, and \protect\uline{underlining} marks nonzero second-best results.}
\label{tab:network_complete_recall}
\footnotesize
\setlength{\tabcolsep}{3.0pt}
\renewcommand{\arraystretch}{1.15}
\begin{tabular*}{\linewidth}{@{\extracolsep{\fill}}ccccc@{}}
\toprule
\makecell[c]{Model}
& \makecell[c]{Sel. Copy.\\Exact}
& \makecell[c]{MAD ICR\\Exact}
& \makecell[c]{MAD Noisy\\Exact}
& \makecell[c]{STORE--RECALL\\Word} \\
\midrule
PMSN \citep{chen2026pmsn} & 0.00 & 0.00 & 0.00 & 63.46 \\
D-RF \citep{zhang2025drf} & 0.00 & \uline{93.13} & \uline{94.22} & \textbf{100.00} \\
SpikingSSMs \citep{shen2025spikingssms} & \textbf{75.66} & 56.80 & 25.55 & \uline{99.82} \\
S5-RF \citep{huber2024scaling} & 0.00 & 0.00 & 0.00 & 67.38 \\
Spiking LMUFormer \citep{liu2024lmuformer} & 0.00 & 0.00 & 0.00 & 0.00 \\
\midrule
TV-MCN & \uline{49.93} & \textbf{99.92} & \textbf{99.69} & \uline{99.82} \\
\bottomrule
\end{tabular*}
\end{table}

\end{document}